%% file: main.tex
\documentclass{article} %
\usepackage{iclr2025_conference,times}

\input{math_commands.tex}

\input{custom.tex}

\title{Stabilizing Reinforcement Learning in \\Differentiable Multiphysics Simulation}

\author{%
Eliot Xing \& Vernon Luk \& Jean Oh\\
Carnegie Mellon University\\
\texttt{\{etaoxing,\:vluk,\:jeanoh\}@cmu.edu}
}

\iclrfinalcopy %
\begin{document}

\maketitle

\begin{abstract}
Recent advances in GPU-based parallel simulation have enabled practitioners to collect large amounts of data and train complex control policies using deep reinforcement learning (RL), on commodity GPUs. However, such successes for RL in robotics have been limited to tasks sufficiently simulated by fast rigid-body dynamics. Simulation techniques for soft bodies are comparatively several orders of magnitude slower, thereby limiting the use of RL due to sample complexity requirements. To address this challenge, this paper presents both a novel RL algorithm and a simulation platform to enable scaling RL on tasks involving rigid bodies and deformables. We introduce \OURALGO (\ouralgo), a maximum entropy first-order model-based actor-critic RL algorithm, which uses first-order analytic gradients from differentiable simulation to train a stochastic actor to maximize expected return and entropy. Alongside our approach, we develop \oursim, a parallel differentiable multiphysics simulation platform that supports simulating various materials beyond rigid bodies. We re-implement challenging manipulation and locomotion tasks in \oursim, and show that \ouralgo outperforms baselines over a range of tasks that involve interaction between rigid bodies, articulations, and deformables.
Additional details at~\href{https://\pagelink}{\texttt{\pagelink}}.
\end{abstract}

\section{Introduction}

Progress in deep reinforcement learning (RL) has produced policies capable of impressive behavior, from playing games with superhuman performance~\citep{silver2016mastering, vinyals2019grandmaster} to controlling robots for assembly~\citep{tang2023industreal}, dexterous manipulation~\citep{andrychowicz2020learning, akkaya2019solving}, navigation~\citep{wijmans2020dd, kaufmann2023champion}, and locomotion~\citep{rudin2021learning, radosavovic2024real}. However, standard model-free RL algorithms are extremely sample inefficient. Thus, the main practical bottleneck when using RL is the cost of acquiring large amounts of training data.

To scale data collection for online RL, prior works developed distributed RL frameworks~\citep{nair2015massively, horgan2018distributed, espeholt2018impala} that run many processes across a large compute cluster, which is inaccessible to most researchers and practitioners. More recently, GPU-based parallel environments~\citep{dalton2020accelerating, freeman2021brax, liang2018gpu, makoviychuk2021isaac, mittal2023orbit, gu2023maniskill} have enabled training RL at scale on a single consumer GPU. 

However, such successes of scaling RL in robotics have been limited to tasks sufficiently simulated by fast rigid-body dynamics~\citep{makoviychuk2021isaac}, while physics-based simulation techniques for soft bodies are comparatively several orders of magnitude slower. Consequently for tasks involving deformable objects, such as robotic manipulation of rope~\citep{nair2017combining, chi2022irp}, cloth~\citep{ha2022flingbot, lin2022learning}, elastics~\citep{shen2022acid}, liquids~\citep{ichnowski2022gomp, zhou2023fluidlab}, dough~\citep{shi2022robocraft, shi2023robocook, lin2023planning}, or granular piles~\citep{wang2023dynamic, xue2023neural}, approaches based on motion planning, trajectory optimization, or model predictive control have been preferred over and outperform RL~\citep{huang2020plasticinelab, chen2022daxbench}.

\begin{figure}[t]
    \centering
    \captionsetup[subfigure]{labelformat=empty,skip=2pt}
    \renewcommand\thesubfigure{}
    \captionsetup[subfigure]{position=above}
    \setlength{\tabcolsep}{0.1em}
    \def\arraystretch{3.5}
    \vspace{-3.0em}
    \begin{tabular}{ccc}
    \subcaptionbox{\myfont\normalsize{AntRun}}{\includegraphics[width=0.32\columnwidth]{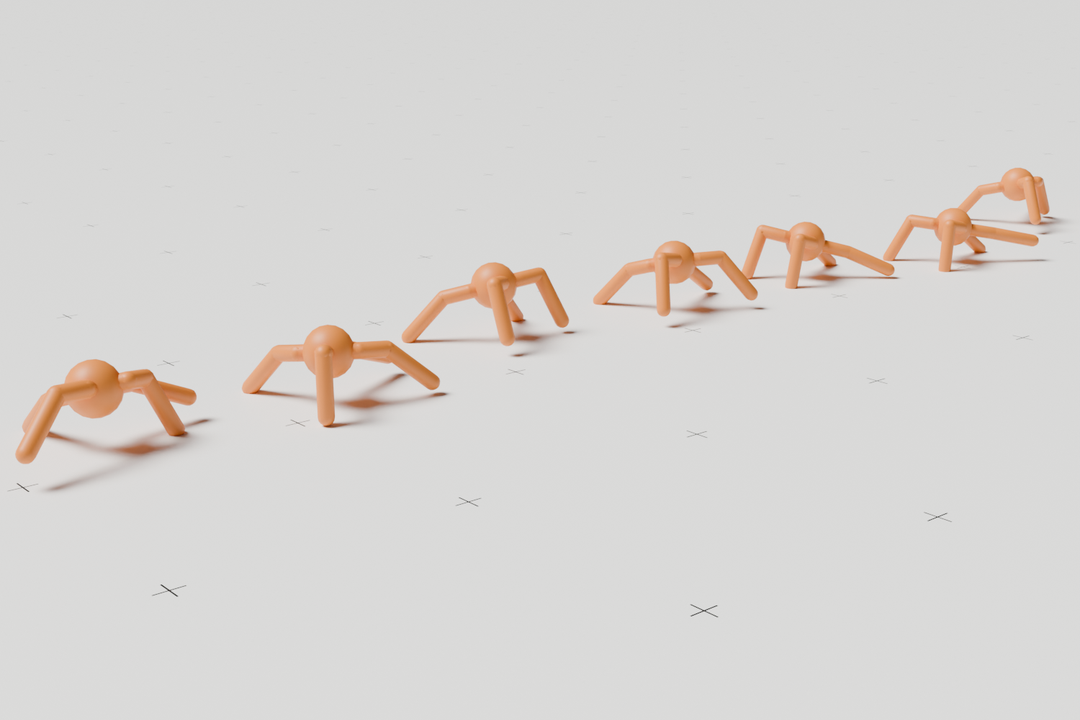}} &
    \subcaptionbox{\myfont\normalsize{HandReorient}}{\includegraphics[width=0.32\columnwidth]{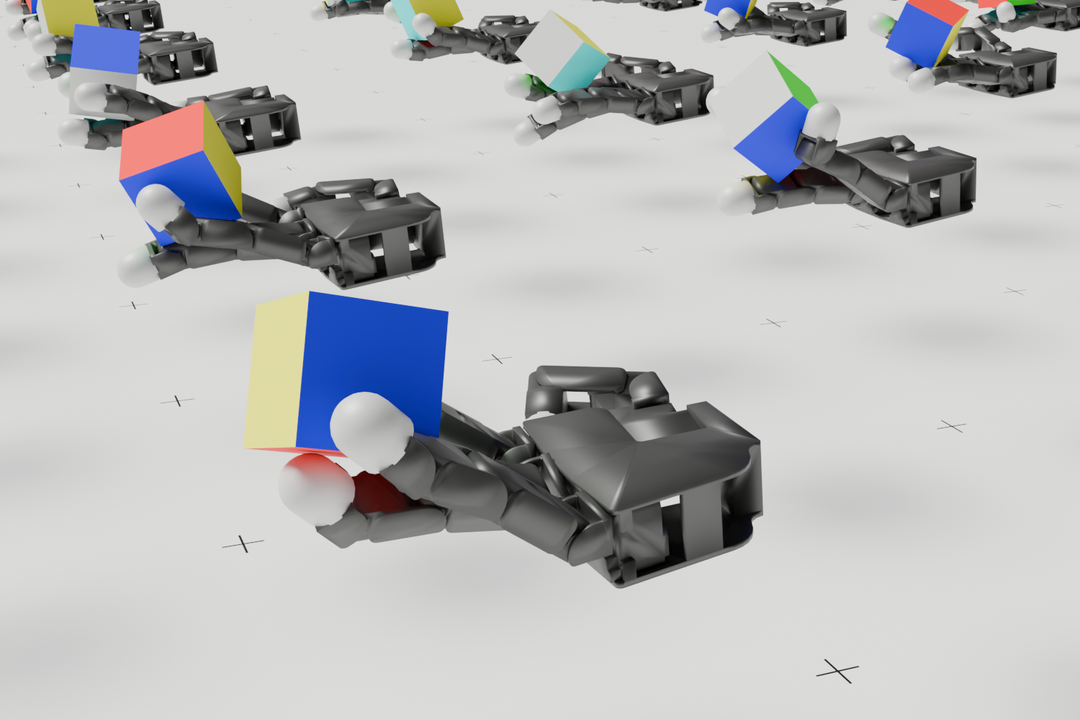}} &
    \subcaptionbox{\myfont\normalsize{RollingFlat}}{\includegraphics[width=0.32\columnwidth]{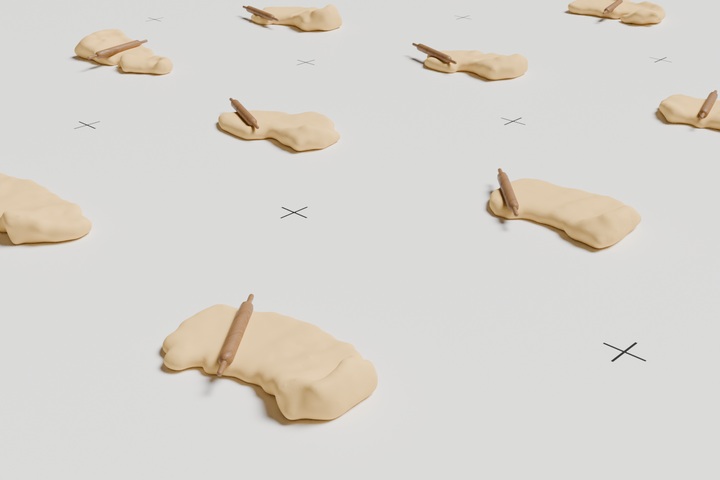}}
    \\
    \subcaptionbox{\myfont\normalsize{SoftJumper}}{\includegraphics[width=0.32\columnwidth]{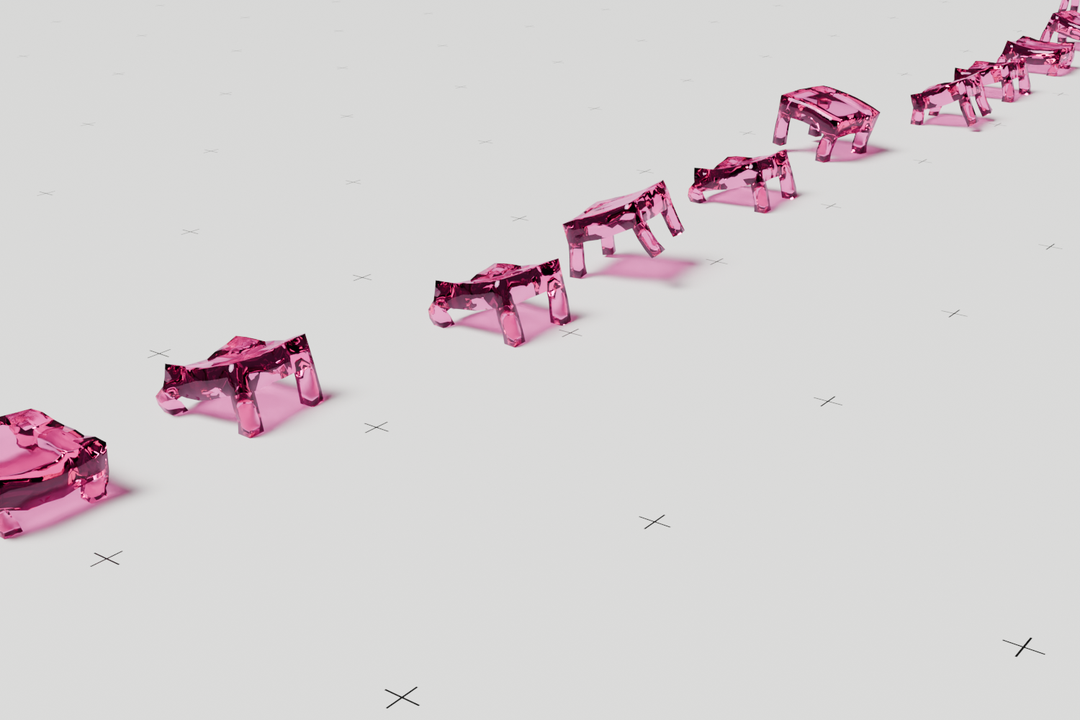}} &
    \subcaptionbox{\myfont\normalsize{HandFlip}}{\includegraphics[width=0.32\columnwidth]{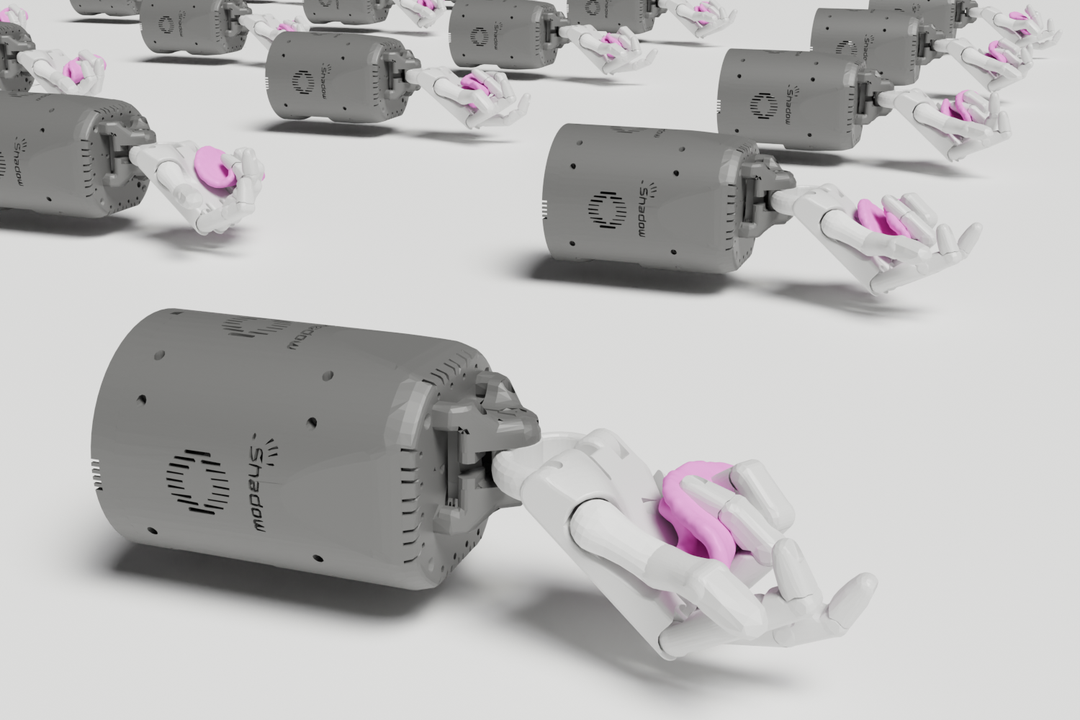}} &
    \subcaptionbox{\myfont\normalsize{FluidMove}}{\includegraphics[width=0.32\columnwidth]{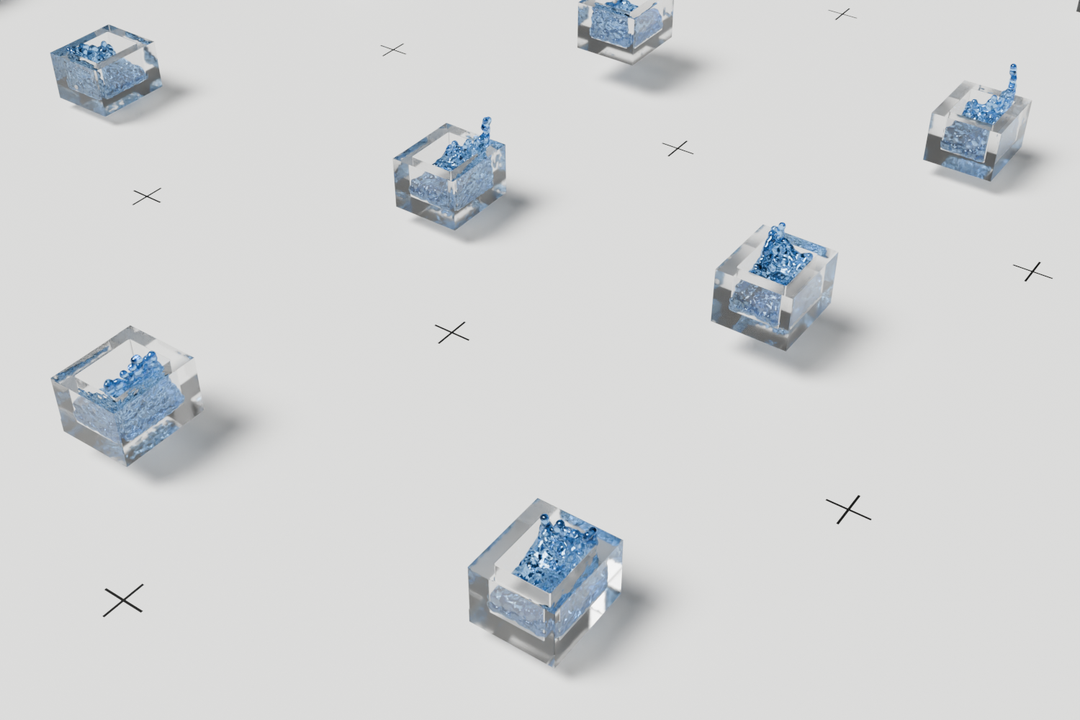}}
    \end{tabular}
    \caption{
        \textbf{Visualizations of tasks implemented in \oursim.} These are manipulation and locomotion tasks involving rigid and soft bodies. AntRun and HandReorient are tasks with articulated rigid bodies, while RollingFlat, SoftJumper, HandFlip, and FluidMove are tasks with deformables.
    }
    \label{fig:oursim_viz}
    \vspace{-1.75em}
\end{figure}

How can we overcome this data bottleneck to scaling RL on tasks involving deformables? Model-based reinforcement learning (MBRL) has shown promise at reducing sample complexity, by leveraging some known model or learning a world model to predict environment dynamics and rewards~\citep{moerland2023model}. In contrast to rigid bodies however, soft bodies have more complex dynamics and higher-dimensional state spaces. This makes learning to model dynamics of deformables highly nontrivial~\citep{lin2021softgym}, often requiring specialized systems architecture and material-specific assumptions such as volume preservation or connectivity.

Recent developments in differentiable physics-based simulators of deformables~\citep{hu2019chainqueen, du2021diffpd, huang2020plasticinelab, zhou2023fluidlab, wang2024thinshell, liang2019differentiable, qiao2021differentiable, li2022diffcloth, heiden2023disect} have shown that first-order gradients from differentiable simulation can be used for gradient-based trajectory optimization and achieve low sample complexity. Yet such approaches are sensitive to initial conditions and get stuck in local optima due to non-smooth optimization landscapes or discontinuities induced by contacts~\citep{li2022contact, antonova2023rethinking}. Additionally, existing soft-body simulations are not easily parallelized, which limits scaling RL in them. Overall, there is no existing simulation platform that is parallelized, differentiable, and supports interaction between articulated rigid bodies and deformables.

In this paper, we approach the sample efficiency problem using first-order model-based RL (FO-MBRL), which leverages first-order analytic gradients from differentiable simulation to accelerate policy learning, without explicitly learning a world model. Thus far, FO-MBRL has been shown to achieve low sample complexity on articulated rigid-body locomotion tasks~\citep{freeman2021brax, xu2021accelerated}, but has not yet been shown to work well for tasks involving deformables~\citep{chen2022daxbench}. We hypothesize that entropy regularization can stabilize policy optimization over analytic gradients from differentiable simulation, such as by smoothing the optimization landscape~\citep{ahmed2019understanding}. To this end, we introduce a novel maximum entropy FO-MBRL algorithm, alongside a parallel differentiable multiphysics simulation platform for RL.

\textbf{Contributions.} \textbf{i)} We introduce \OURALGO (\ouralgo), a first-order MBRL algorithm based on the maximum entropy RL framework. We formulate \ouralgo as an on-policy actor-critic RL algorithm, where a stochastic actor is trained to maximize expected return and entropy using first-order analytic gradients from differentiable simulation. \textbf{ii)} We present \oursim, a scalable and easy-to-use platform which enables parallelizing RL environments of GPU-accelerated differentiable multiphysics simulation and supports various materials beyond rigid bodies. \textbf{iii)} We demonstrate that parallel differentiable simulation enables \ouralgo to outperform baselines over a range of challenging manipulation and locomotion tasks re-implemented using \oursim that involve interaction between rigid bodies, articulations, and deformables such as elastic, plasticine, or fluid materials.

\section{Related Work}

We refer the reader to~\citep{newbury2024review} for an overview of differentiable simulation. We cover \textit{non-parallel} differentiable simulation and model-based RL in Appendix~\ref{sec:extendedrelatedwork}.

\begin{table}[t]
\begin{center}
\makebox[\linewidth][c]{%
\begin{tabular}{lcccccc}
    \toprule
    \multirow{3}{*}{Simulator} & \multirow{3}{*}{\:$\nabla$?\:} & \multicolumn{5}{c}{Materials?} \\
    \cmidrule{3-7}
    & & Rigid & Articulated & Elastic & Plasticine & Fluid \\
    \midrule
    Isaac Gym & \N & \Y & \Y & \Y & \N & \N \\
    Isaac Lab / Orbit & \N & \Y & \Y & \:\:\Y* & \N & \:\:\Y* \\
    ManiSkill & \N & \Y & \Y & \N & \:\:\Y* & \:\:\Y* \\
    \midrule
    TinyDiffSim & \Y & \Y & \Y & \N & \N & \N \\
    Brax & \Y & \Y & \Y & \N & \N & \N \\
    MJX & \:\:\Y* & \Y & \Y & \N & \N & \N \\
    DaXBench & \Y & \Y & \N & \N & \:\:\Y* & \Y \\
    DFlex & \Y & \Y & \Y & \:\:\Y* & \N & \N \\
    \oursim (ours) & \Y & \Y & \Y & \Y & \Y & \Y \\
    \bottomrule
\end{tabular}
}
\end{center}
\caption{\textbf{Comparison of physics-based parallel simulation platforms for RL.} We use * to indicate incomplete feature support at the time of writing. i) \href{https://github.com/isaac-sim/IsaacLab/issues/748}{Isaac Lab / Orbit}: Missing deformable tasks due to breaking API changes and poor simulation stability / scaling. ii) \href{https://github.com/haosulab/ManiSkill/issues/223}{ManiSkill}: The latest version ManiSkill3 does not yet support the soft body tasks introduced in v2. iii) \href{https://github.com/google-deepmind/mujoco/issues/1182}{MJX}: Stability issues with autodifferentiation and gradients. iv) \href{https://github.com/AdaCompNUS/DaXBench/issues/5}{DaXBench}: Plasticine task was omitted from benchmark and requires additional development. v) DFlex: While later work~\citep{murthy2021gradsim, heiden2023disect} has built on DFlex to support elastic and cloth materials, their simulations were not parallelized.}
\label{tab:comparesims}
\vspace{-1.75em}
\end{table}

\textbf{Parallel differentiable simulation.} There are few prior works on parallel differentiable simulators capable of running many environments together, while also computing simulation gradients in batches. TinyDiffSim~\citep{heiden2021neuralsim} implements articulated rigid-body dynamics and contact models in C++/CUDA that can integrate with various autodifferentiation libraries. Brax~\citep{freeman2021brax} implements a parallel simulator in JAX for articulated rigid-body dynamics with simple collision shape primitives. Recently, MJX is building on Brax to provide a JAX re-implementation of MuJoCo~\citep{todorov2012mujoco}, a physics engine widely used in RL and robotics, but does not have feature parity with MuJoCo yet. These aforementioned parallel differentiable simulators are only capable of modeling articulated rigid bodies. DaXBench~\citep{chen2022daxbench} also uses JAX to enable fast parallel simulation of deformables such as rope and liquid by Material Point Method (MPM) or cloth by mass-spring systems, but does not support articulated rigid bodies. DFlex~\citep{xu2021accelerated} presents a differentiable simulator based on source-code transformation~\citep{griewank2008evaluating, hu2020difftaichi} of simulation kernel code to C++/CUDA, that integrates with PyTorch for tape-based autodifferentiation.~\citet{xu2021accelerated} use DFlex for parallel simulation of articulated rigid bodies for high-dimensional locomotion tasks. Later work~\citep{murthy2021gradsim, heiden2023disect} also used DFlex to develop differentiable simulations of cloth and elastic objects, but these were not parallelized and did not support interaction with articulated rigid bodies. To the best of our knowledge, there is no existing differentiable simulation platform that is parallelized with multiphysics support for interaction between rigid bodies, articulations, and various deformables. In this paper, we aim to close this gap with \oursim, our platform for parallel differentiable multiphysics simulation, and in Table~\ref{tab:comparesims} we compare \oursim against existing physics-based parallel simulation platforms.

\textbf{Learning control with differentiable physics.} Gradient-based trajectory optimization is commonly used with differentiable simulation of soft bodies~\citep{hu2019chainqueen, hu2020difftaichi, huang2020plasticinelab, li2023dexdeform, zhou2023fluidlab, wang2024thinshell, si2024difftactile, du2021diffpd, li2022diffcloth, rojas2021differentiable, qiao2020scalable, qiao2021differentiable, liu2023softmac, chen2022daxbench, heiden2023disect}. Differentiable physics can provide physical priors for control in end-to-end learning systems, such as for quadruped locomotion~\citep{song2024learning}, drone navigation~\citep{zhang2024back}, robot painting~\citep{schaldenbrand2023frida}, or motion imitation~\citep{ren2023diffmimic}. Gradients from differentiable simulation can also be directly used for policy optimization. PODS~\citep{mora2021pods} proposes a first and second order policy improvement based on analytic gradients of a value function with respect to the policy's action outputs. APG~\citep{freeman2021brax} uses analytic simulation gradients to directly compute policy gradients. SHAC~\citep{xu2021accelerated} presents an actor-critic algorithm, where the actor is optimized over a short horizon using analytic gradients, and a terminal value function helps smooth the optimization landscape. AHAC~\citep{georgiev2024adaptive} modifies SHAC to adjust the policy horizon by truncating stiff contacts based on contact forces or the norm of the dynamics Jacobian. Several works also propose different ways to overcome bias and non-smooth dynamics resulting from contacts, by reweighting analytic gradients~\citep{gao2024adaptivegradient, son2024gradient} or explicit smoothing~\citep{suh2022differentiable, zhang2023adaptive, schwarke2024learning}. In this work, we propose a maximum entropy FO-MBRL algorithm to stabilize policy learning with gradients from differentiable simulation.

\section{Background}

\textbf{Reinforcement learning} (RL) considers an agent interacting with an environment, formalized as a Markov decision process (MDP) represented by a tuple $(\mathcal{S}, \mathcal{A}, P, R, \rho_0, \gamma)$. In this work, we consider discrete-time, infinite-horizon MDPs with continuous action spaces, where $\bm{s} \in \mathcal{S}$ are states, $\bm{a} \in \mathcal{A}$ are actions, $P: \mathcal{S} \times \mathcal{A} \rightarrow \mathcal{S}$ is the transition function, $R : \mathcal{S} \times \mathcal{A} \rightarrow \mathbb{R}$ is a reward function, $\rho_0(\bm{s})$ is an initial state distribution, and $\gamma$ is the discount factor. We want to obtain a policy $\pi : \mathcal{S} \rightarrow \mathcal{A}$ which maximizes the expected discounted sum of rewards (return) $\Exp_{\pi} [ \sum_{t=0}^{\infty} \gamma^t r_t ]$ with $r_t = R(\bm{s}_t, \bm{a}_t)$, starting from state $\bm{s}_0 \sim \rho_0$. We also denote the state distribution $\rho_{\pi}(\bm{s})$ and state-action distribution $\rho_{\pi}(\bm{s},\bm{a})$ for trajectories generated by a policy $\pi(\bm{a}_t | \bm{s}_t)$.

In practice, the agent interacts with the environment for $T$ steps in a finite-length episode, yielding a trajectory $\tau = (\bm{s}_0, \bm{a}_0, \bm{s}_1, \bm{a}_1, \ldots, \bm{s}_{T-1}, \bm{a}_{T-1})$. We can define the $H$-step return :
\begin{equation}
    R_{0:H} (\tau) = \sum_{t=0}^{H-1} \gamma^t r_t ,
\end{equation}
and standard RL objective to optimize $\theta$ parameterizing a policy $\pi_\theta$ to maximize the expected return :
\begin{equation}
    J(\pi) = \Exp_{\substack{\bm{s}_0 \sim \rho_0\\ \:\:\: \tau \sim \rho_\pi}} [R_{0:T}] .
\label{eq:piloss}
\end{equation}
Typically, the policy gradient theorem~\citep{sutton1999policy} provides a useful expression of $\nabla_\theta J(\pi)$ that does not depend on the derivative of state distribution $\rho_\pi(\cdot)$ :
\begin{equation}
    \nabla_{\theta} J(\pi) \propto \int_{\mathcal{S}} \rho_{\pi}(\bm{s}) \int_{\mathcal{A}} \nabla_{\theta} \pi(\bm{a} | \bm{s}) Q^{\pi} (\bm{s}, \bm{a}) \:d\bm{a} \:d\bm{s} ,
\end{equation}
where $Q^{\pi}(\bm{s}_t, \bm{a}_t) = \Exp_{\tau \sim \rho_\pi} [ R_{t:T} ]$ is the $Q$-function (state-action value function).

We proceed to review zeroth-order versus first-order estimators of the policy gradient following the discussion in~\citep{suh2022differentiable, georgiev2024adaptive}. We denote a single zeroth-order estimate :
\begin{equation}
    \hat{\nabla}^{[0]}_\theta J(\pi) = R_{0:T} \sum_{t=0}^{T-1} \nabla_{\theta} \log \pi (\bm{a}_t | \bm{s}_t) ,
\label{eq:pg}
\end{equation}
where the zeroth-order batched gradient (ZOBG) is the sample mean $\overline{\nabla}^{[0]}_\theta J(\pi)=\frac{1}{N} \sum_{i=1}^N \hat{\nabla}^{[0]}_\theta J(\pi)$ and is an unbiased estimator, under some mild assumptions to ensure the gradients are well-defined. The ZOBG yields an $N$-sample Monte-Carlo estimate commonly known as the REINFORCE estimator~\citep{williams1992simple} in RL literature, or the score function / likelihood-ratio estimator. Policy gradient methods may use different forms of Equation~\ref{eq:pg} to adjust the bias and variance of the estimator~\citep{schulman2015high}. For instance, a baseline term can be used to reduce variance of the estimator, by substituting $R_{0:T}$ with $R_{0:T} - R_{l:H+l}$.

\textbf{Differentiable simulation} as the environment provides gradients for the transition dynamics $P$ and rewards $R$, so we can directly obtain an analytic value for $\nabla_{\theta} R_{0:T}$ under policy $\pi_\theta$. In this setting, for a single first-order estimate :
\begin{equation}
    \hat{\nabla}^{[1]}_\theta J(\pi) = \nabla_\theta R_{0:T} ,
\end{equation}
then the first-order batched gradient (FOBG) is the sample mean $\overline{\nabla}^{[1]}_\theta J(\pi) = \frac{1}{N} \sum_{i=1}^N \hat{\nabla}^{[1]}_\theta J(\pi)$, and is also known as the pathwise derivative~\citep{schulman2015gradient} or reparameterization trick~\citep{kingma2014auto, rezende2014stochastic, titsias2014doubly}.

\textbf{First-order model-based RL} (FO-MBRL) aims to use differentiable simulation (and its first-order analytic gradients) as a known differentiable model, in contrast to vanilla MBRL which either assumes a given non-differentiable model or learns a world model of dynamics and rewards from data.

\textbf{Analytic Policy Gradient} (APG,~\cite{freeman2021brax}) uses FOBG estimates to directly maximize the discounted return over a truncated horizon :
\begin{equation}
    J(\pi) = \sum_{l=t}^{t+H-1} \Exp_{(\bm{s}_l, \bm{a}_l) \sim \rho_\pi} [ \gamma^{l - t} r_l ] ,
\end{equation}
and is also referred to as Backpropagation Through Time (BPTT,~\citet{werbos1990backpropagation, mozer1995focused}), particularly when the horizon is the full episode length~\citep{degrave2019differentiable, huang2020plasticinelab}.

\textbf{Short-Horizon Actor-Critic} (SHAC,~\cite{xu2021accelerated}) is a FO-MBRL algorithm which learns a policy $\pi_\theta$ and (terminal) value function $V_{\psi}$ :
\begin{equation}
    J(\pi) = \sum_{l=t}^{t+H-1} \Exp_{(\bm{s}_l, \bm{a}_l) \sim \rho_\pi} [ \gamma^{l - t} r_l + \gamma^t V(\bm{s}_{t+H}) ] ,
\end{equation}
\begin{equation}
    \mathcal{L}(V) = \sum_{l=t}^{t+H - 1} \Exp_{\bm{s}_l \sim \rho_\pi} [ || V(\bm{s}) - \tilde{V}(\bm{s}) ||^2 ] ,
    \label{eq:shaccriticloss}
\end{equation}
where $\tilde{V}(\bm{s}_t)$ are value estimates for state $\bm{s}_t$ computed starting from time step $t$ over an $H$-step horizon. TD$(\lambda)$~\citep{sutton1988learning} is used for value estimation, which computes $\lambda$-returns $G^\lambda_{t:t+H}$ as a weighted average of value-bootstrapped $k$-step returns $G_{t:t+k}$ :
\begin{equation}
    \tilde{V}(\bm{s}_t) = G^\lambda_{t:t+H} = (1-\lambda) \left( \sum_{l=1}^{H-1-t} \lambda^{l-1} G_{t:t+l} \right) + \lambda^{H-t-1} G_{t:t+H} ,
    \label{eq:tdlambda}
\end{equation}
where $G_{t:t+k}=\left( \sum_{l=0}^{k-1} \gamma^{l} r_{t+l} \right) + \gamma^{k} V (\bm{s}_{t+k})$. The policy and value function are optimized in an alternating fashion per standard actor-critic formulation~\citep{konda1999actor}. The policy gradient is obtained by FOBG estimation, with single first-order estimate :
\begin{equation}
    \hat{\nabla}^{[1]}_\theta J(\pi) = \nabla_\theta (R_{0:H} + \gamma^H V(\bm{s}_H)) ,
\end{equation}
and the value function is optimized as usual by backpropagating $\nabla_{\psi} \mathcal{L}(V)$ of the mean-squared loss in Eq.~\ref{eq:shaccriticloss}. Combining value estimation with a truncated short-horizon window where $H \ll T$~\citep{williams1995gradient}, SHAC optimizes over a smoother surrogate reward landscape compared to BPTT over the entire $T$-step episode.

\section{\OURALGO (\ouralgo)}
\label{sec:algo}

Empirically we observe that SHAC, a state-of-the-art FO-MBRL algorithm, is still prone to suboptimal convergence to local minima in the reward landscape (Appendix, Figure~\ref{fig:shac_individualruns_dflexant4m}). We hypothesize that entropy regularization can stabilize policy optimization over analytic gradients from differentiable simulation, such as by smoothing the optimization landscape~\citep{ahmed2019understanding} or providing robustness under perturbations~\citep{eysenbach2022maximum}.

We draw on the maximum entropy RL framework~\citep{kappen2005path, todorov2006linearly, ziebart2008maximum, toussaint2009robot, theodorou2010generalized, haarnoja2017reinforcement} to formulate \OURALGO (\ouralgo), a maximum entropy FO-MBRL algorithm (Section~\ref{sec:maxent}). To implement \ouralgo, we make several design choices, including modifications building on SHAC (Section~\ref{sec:designchoices}). In Appendix~\ref{sec:visualencoders}, we describe how we use visual encoders to learn policies from high-dimensional visual observations in differentiable simulation. Pseudocode for \ouralgo is shown in Appendix~\ref{sec:pseudocode}, and the computational graph of SAPO is illustrated in Appendix Figure~\ref{fig:algo_diagram}.

\subsection{Maximum entropy RL in differentiable simulation}
\label{sec:maxent}

\textbf{Maximum entropy RL}~\citep{ziebart2008maximum, ziebart2010modeling} augments the standard (undiscounted) return maximization objective with the expected entropy of the policy over $\rho_{\pi} (\bm{s}_t)$ :
\begin{equation}
    J(\pi) = \sum_{t=0}^{\infty} \Exp_{(\bm{s}_t, \bm{a}_t) \sim \rho_\pi} [r_t + \alpha \mathcal{H}_{\pi} [ \bm{a}_t | \bm{s}_t ] ] ,
\end{equation}
where $\mathcal{H}_{\pi}[\bm{a}_t | \bm{s}_t]=-\int_{\mathcal{A}} \pi(\bm{a}_t | \bm{s}_t) \log \pi (\bm{a}_t | \bm{s}_t) d\bm{a}_t$ is the continuous Shannon entropy of the action distribution, and the temperature $\alpha$ balances the entropy term versus the reward. 

Incorporating the discount factor~\citep{thomas2014bias, haarnoja2017reinforcement}, we obtain the following objective which maximizes the expected return and entropy for future states starting from $(\bm{s}_t, \bm{a}_t)$ weighted by its probability $\rho_\pi$ under policy $\pi$ :
\begin{equation}
    J_{\mathrm{maxent}}(\pi) = \sum_{t=0}^{\infty} \Exp_{(\bm{s}_t, \bm{a}_t) \sim \rho_\pi} \left[ \sum_{l=t}^{\infty} \gamma^{l-t} \Exp_{(\bm{s}_l, \bm{a}_l)\sim \rho_{\pi}} [r_t + \alpha \mathcal{H}_\pi [ \bm{a}_l | \bm{s}_l ] ] \right] .
    \label{eq:fullmaxentcost}
\end{equation}
The soft $Q$-function is the expected value under $\pi$ of the discounted sum of rewards and entropy :
\begin{equation}
    Q^{\pi}_{\mathrm{soft}} (\bm{s}_t, \bm{a}_t) = r_t + \Exp_{(\bm{s}_{t+1}, \ldots)\sim \rho_\pi} \left[ \sum_{l=t+1}^\infty \gamma^l ( r_{l} + \alpha \mathcal{H}_\pi [ \bm{a}_l | \bm{s}_l ] ) \right] ,
\end{equation}
and the soft value function is :
\begin{equation}
    V^{\pi}_{\mathrm{soft}} (\bm{s}_t) = \alpha \log \int_{\mathcal{A}} \exp ( \frac{1}{\alpha} Q^{\pi}_{\mathrm{soft}} (\bm{s}, \bm{a}) ) d\bm{a} .
\end{equation}
When $\pi(\bm{a} | \bm{s}) = \exp(\frac{1}{\alpha}(Q^{\pi}_{\mathrm{soft}}(\bm{s}, \bm{a})  - V^{\pi}_{\mathrm{soft}}(\bm{s}))) \triangleq \pi^*$, then the soft Bellman equation yields the following relationship :
\begin{equation}
    Q^{\pi}_{\mathrm{soft}} (\bm{s}_t, \bm{a}_t) = r_t + \gamma \Exp_{(\bm{s}_{t+1}, \ldots)\sim \rho_\pi} [ V^{\pi}_{\mathrm{soft}} (\bm{s}_{t+1}) ] ,
\end{equation}
where we can rewrite the discounted maximum entropy objective in Eq.~\ref{eq:fullmaxentcost} :
\begin{align}
    J_{\mathrm{maxent}}(\pi) &= \sum_{t=0}^{\infty} \Exp_{(\bm{s}_t, \bm{a}_t) \sim \rho_\pi} \left[ Q^{\pi}_{\mathrm{soft}} (\bm{s}, \bm{a}) + \alpha \mathcal{H}_\pi [ \bm{a}_t | \bm{s}_t ]  \right]
    \\
    &= \sum_{t=0}^{\infty} \Exp_{(\bm{s}_t, \bm{a}_t) \sim \rho_\pi} \left[r_t + \alpha \mathcal{H}_\pi [ \bm{a}_t | \bm{s}_t ] + \gamma  V^{\pi}_{\mathrm{soft}} (\bm{s}_{t+1})  \right] .
    \label{eq:maxentcostsoftval}
\end{align}
By Soft Policy Iteration~\citep{haarnoja2018soft}, the soft Bellman operator $\mathcal{B}^*$ defined by $(\mathcal{B}^* Q)(\bm{s}_t, \bm{a}_t) = r_t + \gamma \Exp_{\bm{s}_{t+1}\sim \rho_\pi}[ V(\bm{s}_{t+1}) ]$ has a unique contraction $Q^* = \mathcal{B}^* Q^*$~\citep{fox2016taming} and converges to the optimal policy $\pi^*$.

\textbf{Our main observation} is when the environment is a differentiable simulation, we can use FOBG estimates to directly optimize $J_{\mathrm{maxent}}(\pi)$, including discounted policy entropy. Consider the entropy-augmented $H$-step return :
\begin{equation}
    R^{\alpha}_{0:H} (\tau) = \sum_{t=0}^{H-1} \gamma^t (r_t + \alpha \mathcal{H}_\pi [ \bm{a}_t | \bm{s}_t ] ) ,
\end{equation}
then we have a single first-order estimate of Eq.~\ref{eq:maxentcostsoftval} :
\begin{equation}
    \hat{\nabla}^{[1]}_\theta J_{\mathrm{maxent}}(\pi) = \nabla_\theta (R^{\alpha}_{0:H} + \gamma^H V_{\mathrm{soft}} (\bm{s}_{H})) .
    \label{eq:soft_fo_estimate}
\end{equation}
Furthermore, we can incorporate the entropy-augmented return into $TD(\lambda)$ estimates of Eq.~\ref{eq:tdlambda} using soft value-bootstrapped $k$-step returns :
\begin{equation}
    \Gamma_{t:t+k}=\left( \sum_{l=0}^{k-1} \gamma^{l} (r_{t+l} + \alpha \mathcal{H}_\pi [ \bm{a}_{t+l} | \bm{s}_{t+l} ]) \right) + \gamma^{k} V_{\mathrm{soft}} (\bm{s}_{t+k}) ,
    \label{eq:soft_bootstrap_ret}
\end{equation}
where $\tilde{V}_{\mathrm{soft}}(\bm{s}_t) = \Gamma^\lambda_{t:t+H}$, and the value function is trained by minimizing Eq.~\ref{eq:shaccriticloss} with $V_{\mathrm{soft}}$, $\tilde{V}_{\mathrm{soft}}$, and $\Gamma_{t:t+k}$ substituted in place of $V$, $\tilde{V}$, and $G_{t:t+k}$. We refer to this maximum entropy FO-MBRL formulation as \textbf{\OURALGO} (\ouralgo). 

Note that we instantiate \ouralgo as an actor-critic algorithm that learns the soft value function by TD learning with on-policy data. In comparison, Soft Actor-Critic (SAC), a popular off-policy maximum entropy model-free RL algorithm, estimates soft $Q$-values by minimizing the soft Bellman residual with data sampled from a replay buffer. Connections may also drawn between \ouralgo to a maximum entropy variant of SVG($H$)~\citep{heess2015learning, amos2021model}, which uses rollouts from a learned world model instead of trajectories from differentiable simulation.

\subsection{Design choices}
\label{sec:designchoices}

\textbf{I. Entropy adjustment.} In practice, we apply automatic temperature tuning~\citep{haarnoja2018softapp} to match a target entropy $\bar{\mathcal{H}}$ via an additional Lagrange dual optimization step :
\begin{equation}
    \min_{\alpha_t \geq 0} \Exp_{(\bm{s}_{t}, \bm{a}_{t})\sim \rho_\pi} [\alpha_t (\mathcal{H}_\pi [ \bm{a}_t | \bm{s}_t ] - \bar{\mathcal{H}})].
    \label{eq:autoent}
\end{equation}
We use $\bar{\mathcal{H}}=-\mathrm{dim}(\mathcal{A}) / 2$ following~\citep{ball2023efficient}.

\textbf{II. Target entropy normalization.} To mitigate non-stationarity in target values~\citep{yu2022you} and improve robustness across tasks with varying reward scales and action dimensions, we normalize entropy estimates. The continuous Shannon entropy is not scale invariant~\citep{marsh2013introduction}. In particular, we offset~\citep{han2021max} and scale entropy by $\bar{\mathcal{H}}$ to be approximately contained within $[0, +1]$.

\textbf{III. Stochastic policy parameterization.} We use state-\textit{dependent} variance, with squashed Normal distribution $\pi_\theta = \mathrm{tanh}(\mathcal{N}(\mu_\theta(\mathbf{s}), \sigma^2_\theta (\mathbf{s})))$, which aligns with SAC~\citep{haarnoja2018softapp}. This enables policy entropy adjustment and captures aleatoric uncertainty in the environment~\citep{kendall2017uncertainties, chua2018deep}. In contrast, SHAC uses state-independent variance, similar to the original PPO implementation~\citep{schulman2017proximal}.

\textbf{IV. Critic ensemble, no target networks.} We use the clipped double critic trick~\citep{fujimoto2018addressing} and also remove the critic target network in SHAC, similar to~\citep{georgiev2024adaptive}. However when updating the actor, we instead compute the \textit{average} over the two value estimates to include in the return (Eq.~\ref{eq:soft_fo_estimate}), while using the \textit{minimum} to estimate target values in standard fashion, following~\citep{ball2023efficient}. While originally intended to mitigate overestimation bias in $Q$-learning (due to function approximation and stochastic optimization~\citep{thrun2014issues}), prior work has shown that the value lower bound obtained by clipping can be overly conservative and cause the policy to pessimistically underexplore~\citep{ciosek2019better, moskovitz2021tactical}.

Target networks~\citep{mnih2015human} are widely used~\citep{lillicrap2015continuous, fujimoto2018addressing, haarnoja2018softapp} to stabilize temporal difference (TD) learning, at the cost of slower training. Efforts have been made to eliminate target networks~\citep{kim2019deepmellow, yang2021overcoming, shao2022grac, gallici2024simplifying}, and recently Cross$Q$~\citep{bhatt2024crossq} has shown that careful use of normalization layers can stabilize off-policy model-free RL to enable removing target networks for improved sample efficiency. Cross$Q$ also reduces Adam $\beta_1$ momentum from $0.9$ to $0.5$, while keeping the default $\beta_2=0.999$. In comparison, SHAC uses $\beta_1=0.7$ and $\beta_2=0.95$. Using smaller momentum parameters decreases exponential decay (for the moving average estimates of the 1st and 2nd moments of the gradient) and effectively gives higher weight to more recent gradients, with less smoothing by past gradient history~\citep{kingma15adam}. 

\textbf{V. Architecture and optimization.} We use SiLU~\citep{elfwing2018silu} instead of ELU for the activation function. We also switch the optimizer from Adam to AdamW~\citep{loshchilov2017decoupled}, and lower gradient norm clipping from $1.0$ to $0.5$. Note that SHAC already uses LayerNorm~\citep{ba2016layernorm}, which has been shown to stabilize TD learning when not using target networks or replay buffers~\citep{bhatt2024crossq, gallici2024simplifying}.

\section{\oursim: Parallel Differentiable Multiphysics Simulation}
\label{sec:sim}

We aim to evaluate our approach on more challenging manipulation and locomotion tasks that involve interaction between articulated rigid bodies and deformables. To this end, we introduce \oursim, our parallel differentiable multiphysics simulation platform that provides GPU-accelerated parallel environments for RL and enables computing batched simulation gradients efficiently. We build \oursim on NVIDIA Warp~\citep{warp2022}, the successor to DFlex~\citep{xu2021accelerated, murthy2021gradsim, turpin2022grasp, heiden2023disect}. 

We proceed to discuss high-level implementation details and optimization tricks to enable efficient parallel differentiable simulation. 
We develop a parallelized implementation of Material Point Method (MPM) which supports simulating parallel environments of complex deformable materials, building on the MLS-MPM implementation by~\citep{ma2023learning} used for non-parallel simulation. Furthermore, we support one-way coupling from kinematic articulated rigid bodies to MPM particles, based on the (non-parallel) MPM-based simulation from~\citep{huang2020plasticinelab, li2023dexdeform}.

\subsection{Parallel differentiable simulation}

We implement all simulation code in NVIDIA Warp~\citep{warp2022}, a library for differentiable programming that converts Python code into CUDA kernels by runtime JIT compilation. Warp implements reverse-mode auto-differentiation through the discrete adjoint method, using a tape to record kernel calls for the computation graph, and generates kernel adjoints to compute the backward pass. Warp uses source-code transformation~\citep{griewank2008evaluating, hu2020difftaichi} to automatically generate kernel adjoints.

We use gradient checkpointing~\citep{griewank2000algorithm, qiao2021efficient} to reduce memory requirements. During backpropogation, we run the simulation forward pass again to recompute intermediate values, instead of saving them during the initial forward pass. This is implemented by capturing and replaying CUDA graphs, for both the forward pass and the backward pass of the simulator. Gradient checkpointing by CUDA graphs enables us to compute batched simulation gradients over multiple time steps efficiently, when using more simulation substeps for simulation stability. We use a custom PyTorch autograd function to interface simulation data and model parameters between Warp and PyTorch while maintaining auto-differentiation functionality.

\section{Experiments}
\label{sec:experiments}

We evaluate our proposed maximum entropy FO-MBRL algorithm, \OURALGO (\ouralgo, Section~\ref{sec:algo}), against baselines on a range of locomotion and manipulation tasks involving rigid and soft bodies. We implement these tasks in \oursim (Section~\ref{sec:sim}), our parallel differentiable multiphysics simulation platform. We also compare algorithms on DFlex rigid-body locomotion tasks introduced in~\citep{xu2021accelerated} in Appendix~\ref{sec:dflexlocomotion}.

\textbf{Baselines.} We compare to vanilla model-free RL algorithms: Proximal Policy Optimization (PPO,~\citet{schulman2017proximal}), an on-policy actor-critic algorithm; Soft Actor-Critic (SAC,~\citet{haarnoja2018softapp}) an off-policy maximum entropy actor-critic algorithm. We use the implementations and hyperparameters from~\citep{li2023parallel} for both, which have been validated to scale well with parallel simulation. Implementation details (network architecture, common hyperparameters, etc.) are standardized between methods for fair comparison, see Appendix~\ref{sec:hyperparams}. We also compare against Analytic Policy Gradient (APG,~\citet{freeman2021brax}) and Short-Horizon Actor-Critic (SHAC,~\citet{xu2021accelerated}), both of which are state-of-the-art FO-MBRL algorithms that leverage first-order analytic gradients from differentiable simulation for policy learning. Finally, we include gradient-based trajectory optimization (TrajOpt) as a baseline, which uses differentiable simulation gradients to optimize for an open-loop action sequence that maximizes total rewards across environments. 

\textbf{Tasks}. Using \oursim, we re-implement a range of challenging manipulation and locomotion tasks involving rigid and soft bodies that have appeared in prior works. \oursim enables training algorithms on parallel environments, and differentiable simulation to compute analytic simulation gradients through environment dynamics and rewards.
We visualize these tasks in Figure~\ref{fig:oursim_viz}. To simulate deformables, we use $\sim2500$ particles per env. See Appendix~\ref{sec:tasks} for more details.
\begin{enumerate}[leftmargin=1em,topsep=0pt,itemsep=0pt,label={}]
    \item \textbf{AntRun} -- Ant locomotion task from DFlex~\citep{xu2021accelerated}, where the objective is to maximize the forward velocity of a four-legged ant rigid-body articulation.

    \item \textbf{HandReorient} -- Allegro hand manipulation task from Isaac Gym~\citep{makoviychuk2021isaac}, where the objective is to perform in-hand dexterous manipulation to rotate a rigid cube given a target pose. We replace non-differentiable terms of the reward function (ie. boolean comparisons) with differentiable alternatives to enable computing analytic gradients.

    \item \textbf{RollingFlat} -- Rolling pin manipulation task from PlasticineLab~\citep{huang2020plasticinelab}, where the objective is to flatten a rectangular piece of dough using a cylindrical rolling pin.

    \item \textbf{SoftJumper} -- Soft jumping locomotion task, inspired by GradSim~\citep{murthy2021gradsim} and DiffTaichi~\citep{hu2020difftaichi}, where the objective is to maximize the forward velocity and height of a high-dimensional actuated soft elastic quadruped.

    \item \textbf{HandFlip} -- Shadow hand flip task from DexDeform~\citep{li2023dexdeform}, where the objective is to flip a cylindrical piece of dough in half within the palm of a dexterous robot hand.

    \item \textbf{FluidMove} -- Fluid transport task from SoftGym~\citep{lin2021softgym}, where the objective is to move a container filled with fluid to a given target position, without spilling fluid out of the container.

\end{enumerate}

Note that $\{$AntRun, HandReorient$\}$ are tasks that involve articulated rigid bodies only, with state-based observations. In contrast, $\{$RollingFlat, SoftJumper, HandFlip, FluidMove$\}$ are tasks that also involve deformables, with both state-based and high-dimensional (particle-based) visual observations.

\begin{figure}[h]
    \centering
    \includegraphics[width=1.0\columnwidth,trim={0 0 4cm 0},clip]{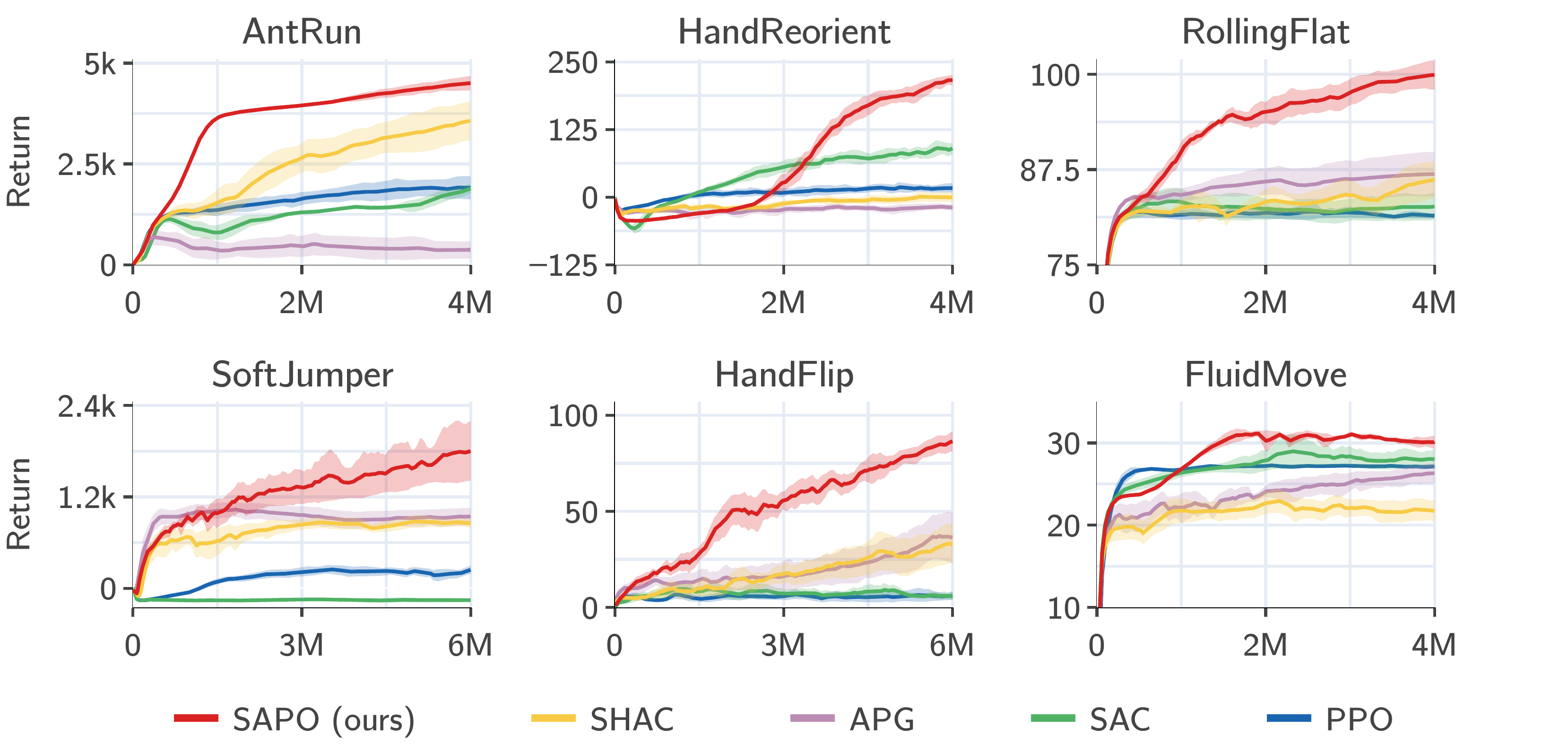}
    \caption{
        \textbf{\oursim tasks training curves.} 
        Episode return as a function of environment steps in \oursim AntRun ($\mathcal{A}\subset\mathbb{R}^{8}$), HandReorient ($\mathcal{A}\subset\mathbb{R}^{16}$), RollingFlat ($\mathcal{A}\subset\mathbb{R}^{3}$), SoftJumper ($\mathcal{A}\subset\mathbb{R}^{222}$), HandFlip ($\mathcal{A}\subset\mathbb{R}^{24}$), and FluidMove ($\mathcal{A}\subset\mathbb{R}^{3}$) tasks. Smoothed using EWMA with $\alpha=0.99$.  Mean and 95\% CIs over 10 random seeds.
    }
    \label{fig:oursim_training}
    \vspace{-1.0em}
\end{figure}

\begin{table}[h]
\begin{center}
\setlength{\tabcolsep}{0.2em}
\begin{tabular}{lcccccc}
    \toprule
     & \textbf{AntRun} & \textbf{HandReorient} & \textbf{RollingFlat} & \textbf{SoftJumper} & \textbf{HandFlip} & \textbf{FluidMove} \\
    \midrule
    PPO & 2048.7 $\pm$ 36.6 & 5.9 $\pm$ 4.9 & 81.2 $\pm$ 0.1 & 261.5 $\pm$ 12.4 & 7.3 $\pm$ 1.1 & 27.3 $\pm$ 0.2 \\
    SAC & 2063.6 $\pm$ 13.9 & 70.5 $\pm$ 10.2 & 83.0 $\pm$ 0.3 & -161.8 $\pm$ 2.5 & 4.6 $\pm$ 1.1 & 28.2 $\pm$ 0.7 \\
    \midrule
    TrajOpt & 915.5 $\pm$ 29.6 & -12.5 $\pm$ 2.0 & 81.5 $\pm$ 0.1 & 437.2 $\pm$ 17.7 & 27.3 $\pm$ 2.6 & 27.0 $\pm$ 0.1 \\
    APG & 258.7 $\pm$ 20.3 & -11.6 $\pm$ 1.9 & 86.9 $\pm$ 0.4 & 956.6 $\pm$ 15.6 & 38.2 $\pm$ 3.5 & 26.3 $\pm$ 0.3 \\
    SHAC & 3621.0 $\pm$ 54.4 & -2.5 $\pm$ 1.8 & 86.8 $\pm$ 0.4 & 853.3 $\pm$ 10.2 & 32.7 $\pm$ 2.9 & 21.7 $\pm$ 0.4 \\
    \rowcolor{rowblue!10}
    \ouralgo (ours) & 4535.9 $\pm$ 24.5 & 221.7 $\pm$ 9.5 & 100.4 $\pm$ 0.4 & 1820.5 $\pm$ 47.9 & 90.0 $\pm$ 2.2 & 30.6 $\pm$ 0.4 \\
    \bottomrule
\end{tabular}
\end{center}
\caption{\textbf{\oursim tasks tabular results.} Evaluation episode returns for final policies after training. Mean and 95\% CIs over 10 random seeds with $2N$ episodes per seed for $N=32$ or $64$ parallel envs.}
\label{tab:oursim_tabular}
\vspace{-1.0em}
\end{table}

\subsection{Results on \oursim tasks}

We compare \ouralgo, our proposed maximum entropy FO-MBRL algorithm, against baselines on a range of challenging manipulation and locomotion tasks that involve rigid and soft bodies, re-implemented in \oursim, our parallel differentiable multiphysics simulation platform. In Figure~\ref{fig:oursim_training}, we visualize training curves to compare algorithms. \ouralgo shows better training stability across different random seeds, against existing FO-MBRL algorithms APG and SHAC. In Table~\ref{tab:oursim_tabular}, we report evaluation performance for final policies after training. \ouralgo outperforms all baselines across all tasks we evaluated, given the same budget of total number of environment steps. We also find that on tasks involving deformables, APG outperforms SHAC, which is consistent with results in DaXBench~\citep{chen2022daxbench} on their set of soft-body manipulation tasks. However, SHAC outperforms APG on the articulated rigid-body tasks, which agrees with the rigid-body locomotion results in DFlex~\citep{xu2021accelerated} that we also reproduce ourselves in Appendix~\ref{sec:dflexlocomotion}. 

In Appendix Figure~\ref{fig:trajviz}, we visualize different trajectories produced by \ouralgo policies after training. We observe that \ouralgo learns to perform tasks with deformables that we evaluate on. For RollingFlat, \ouralgo controls the rolling pin to flatten the dough and spread it across the ground. For SoftJumper, SAPO learns a locomotion policy that controls a soft elastic quadruped to jump forwards. For HandFlip, \ouralgo is capable of controlling a high degree-of-freedom dexterous robot hand, to flip a piece of dough in half within the palm of the hand. For FluidMove, \ouralgo learns a policy to move the container of fluid with minimal spilling. Additionally, \ouralgo learns a successful locomotion policy for the articulated rigid-body locomotion task AntRun. For HandReorient however, \ouralgo is only capable of catching the cube and preventing it from falling to the ground. This is a challenging task that will likely require more environment steps to learn policies capable of re-orienting the cube to given target poses in succession. 

\subsection{\ouralgo ablations}

We investigate which components of \ouralgo yield performance gains over SHAC, on the HandFlip task. We conduct several ablations on \ouralgo: (a) w/o $V_{\mathrm{soft}}$, where instead the critic is trained in standard fashion without entropy in target values; (b) w/o $\mathcal{H}_{\pi}$, where we do not use entropy-augmented returns and instead train the actor to maximize expected returns only; (c) w/o $\mathcal{H}_{\pi}$ and $V_{\mathrm{soft}}$, which corresponds to modifying SHAC by applying design choices $\{$III, IV, V$\}$ described in Section~\ref{sec:designchoices}. 

\begin{figure}[H]
    \begin{minipage}{0.52\columnwidth}
        \centering
        \includegraphics[width=1.0\textwidth,trim={0 0 3cm 0},clip]{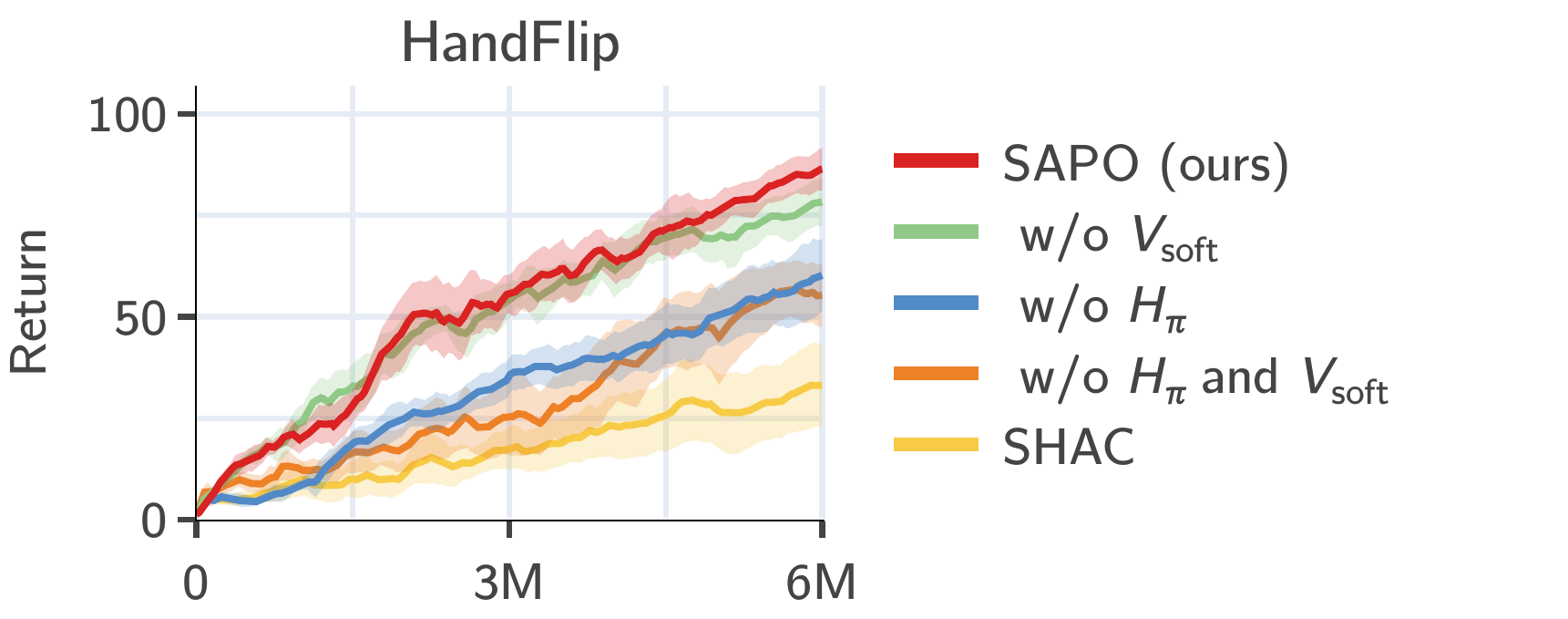}
        \captionof{figure}{
            \textbf{\ouralgo ablations -- HandFlip training curves.} Episode return as a function of environment steps. Smoothed using EWMA with $\alpha=0.99$. Mean and 95\% CIs over 10 random seeds.
        }
        \label{fig:ablations}
    \end{minipage}%
    \hspace{0.5em}
    \begin{minipage}{0.46\columnwidth}
        \centering
        \begin{table}[H]
        \begin{center}
        \setlength{\tabcolsep}{0.25em}
        \vspace{-0.9em}
        \begin{tabular}{lcc}
            \toprule
             & \textbf{HandFlip} & ($\Delta$\%) \\
            \midrule
            \ouralgo (ours) & 90 $\pm$ 2 & +172.7\% \\
            \:\:w/o $V_{\mathrm{soft}}$ & 77 $\pm$ 3 & +133.3\% \\
            \:\:w/o $\mathcal{H}_{\pi}$ & 59 $\pm$ 4 & \:\:+78.8\% \\
            \:\:w/o $\mathcal{H}_{\pi}$ and $V_{\mathrm{soft}}$ & 56 $\pm$ 3 & \:\:+69.7\% \\
            \midrule
            SHAC & 33 $\pm$ 3 & -- \\
            \bottomrule
        \end{tabular}
        \end{center}
        \captionof{table}{\textbf{\ouralgo ablations -- HandFlip tabular results.} Evaluation episode returns for final policies after training. Mean and 95\% CIs over 10 random seeds with 64 episodes per seed.}
        \label{tab:ablation}
        \end{table}
    \end{minipage}
\end{figure}

We visualize training curves in Figure~\ref{fig:ablations}, and in Table~\ref{tab:ablation} we report final evaluation performance as well as percentage change from SHAC's performance as the baseline. From ablation (b), we find that using analytic gradients to train a policy to maximize both expected return and entropy is critical to the performance of \ouralgo, compared to ablation (a) which only replaces the soft value function. 

Additionally, we observe that ablation (c), where we apply design choices $\{$III, IV, V$\}$ onto SHAC, result in approximately half of the performance improvement of \ouralgo over SHAC on the HandFlip task. We also conducted this ablation on the DFlex rigid-body locomotion tasks however, and found these modifications to SHAC to have minimal impact in those settings, shown in Appendix~\ref{sec:ablation_dflex}. We also conduct individual ablations for these three design choices in Appendix~\ref{sec:ablation_designchoices}.

\vspace{-2.0em}
\section{Conclusion}

Due to high sample complexity requirements and slower runtimes for soft-body simulation, RL has had limited success on tasks involving deformables. To address this, we introduce \OURALGO (\ouralgo), a first-order model-based actor-critic RL algorithm based on the maximum entropy RL framework, which leverages first-order analytic gradients from differentiable simulation to achieve higher sample efficiency. Alongside this approach, we present \oursim, a scalable and easy-to-use platform which enables parallelizing RL environments of GPU-based differentiable multiphysics simulation. We re-implement challenging locomotion and manipulation tasks involving rigid bodies, articulations, and deformables using \oursim. On these tasks, we demonstrate that \ouralgo outperforms baselines in terms of sample efficiency as well as task performance given the same budget for total environment steps.

\textbf{Limitations.} \ouralgo relies on end-to-end learning using first-order analytic gradients from differentiable simulation. Currently, we use (non-occluded) subsampled particle states from simulation as observations to policies, which is infeasible to obtain in real-world settings. Future work may use differentiable rendering to provide more realistic visual observations for policies while maintaining differentiability, towards sim2real transfer of policies learned using SAPO. 
Another promising direction to consider is applications between differentiable simulation and learned world models.

\clearpage

\subsubsection*{Acknowledgments}

The authors would like to thank Uksang Yoo, Ananye Agarwal, and David Held for helpful discussions on this work. We would also like to thank Miles Macklin, Eric Heiden, and the rest of the NVIDIA Warp team. This work was in part supported by the Technology Innovation Program (20018112, Development of autonomous manipulation and gripping technology using imitation learning based on visual-tactile sensing) funded by the Ministry of Trade, Industry \& Energy (MOTIE), Korea.

\bibliographystyle{iclr2025_conference}
\setlength{\bibsep}{0.55em}
{
\small
\bibliography{refs}
}

\clearpage

\appendix

\section{Extended Related Work}
\label{sec:extendedrelatedwork}

\textbf{Differentiable simulators.} We review prior works on \textit{non-parallel} differentiable simulators for robotics and control problems.~\citet{degrave2019differentiable} demonstrate that gradient-based optimization can be used to train neural network controllers by implementing a rigid-body physics engine in a modern auto-differentiation library to compute analytic gradients.~\citet{de2018end} show how to analytically differentiate through 2D rigid-body dynamics, by defining a linear complementarity problem (LCP) and using implicit differentiation. Exploiting sparsity to differentiate through the LCP more efficiently, Nimble~\citep{werling2021nimble} presents a differentiable rigid-body physics engine that supports complex 3D contact geometries. Dojo~\citep{howell2022dojo} implements a differentiable rigid-body physics engine with a hard contact model (based on nonlinear friction cones) via a nonlinear complementarity problem (NCP), and a custom interior-point solver to reliably solve the NCP. Also using implicit differentiation,~\citet{qiao2020scalable} adopt meshes as object representations to support both rigid and deformable objects and adopt a sparse local collision handling scheme that enables coupling between objects of different materials.~\citet{geilinger2020add} presents a unified frictional contact model for rigid and deformable objects, while analytically computing gradients through adjoint sensitivity analysis instead. DiffTaichi~\citep{hu2019taichi, hu2020difftaichi}, used within several differentiable soft-body simulators~\citep{hu2019chainqueen, huang2020plasticinelab, zhou2023fluidlab, si2024difftactile, wang2024thinshell}, introduces a differentiable programming framework for tape-based auto-differentiation using source-code transformation to generate kernel adjoints. We build \oursim using NVIDIA Warp~\citep{warp2022}, a differentiable programming framework similar to DiffTaichi.

\textbf{Model-based reinforcement learning.} Generally, approaches in MBRL~\citep{moerland2023model} either assume a known (non-differentiable) model or learn a world model by predicting dynamics and rewards from data. We will focus on the latter, as learned models can be differentiated through for policy learning, similar to FO-MBRL which computes analytic policy gradients by backpropagating rewards from some known differentiable model (ie. differentiable simulator). Value gradient methods estimate policy gradients by directly backpropagating through the learned model~\citep{deisenroth2013gaussian, heess2015learning, feinberg2018model, parmas2018pipps, clavera2020modelaugmented, hafner2020dream}. Learned models can also be used for data augmentation by generating imagined trajectories from the learned model, as first proposed in DYNA~\citep{sutton1990integrated}. In this paper, we consider FO-MBRL~\citep{freeman2021brax, xu2021accelerated} which does analytic policy optimization with data and gradients from a differentiable simulator, without explicitly learning a world model.

\begin{figure}[H]
    \centering
    \includegraphics[width=0.9\columnwidth, trim={0.5cm 0 2cm 0.75cm},clip]{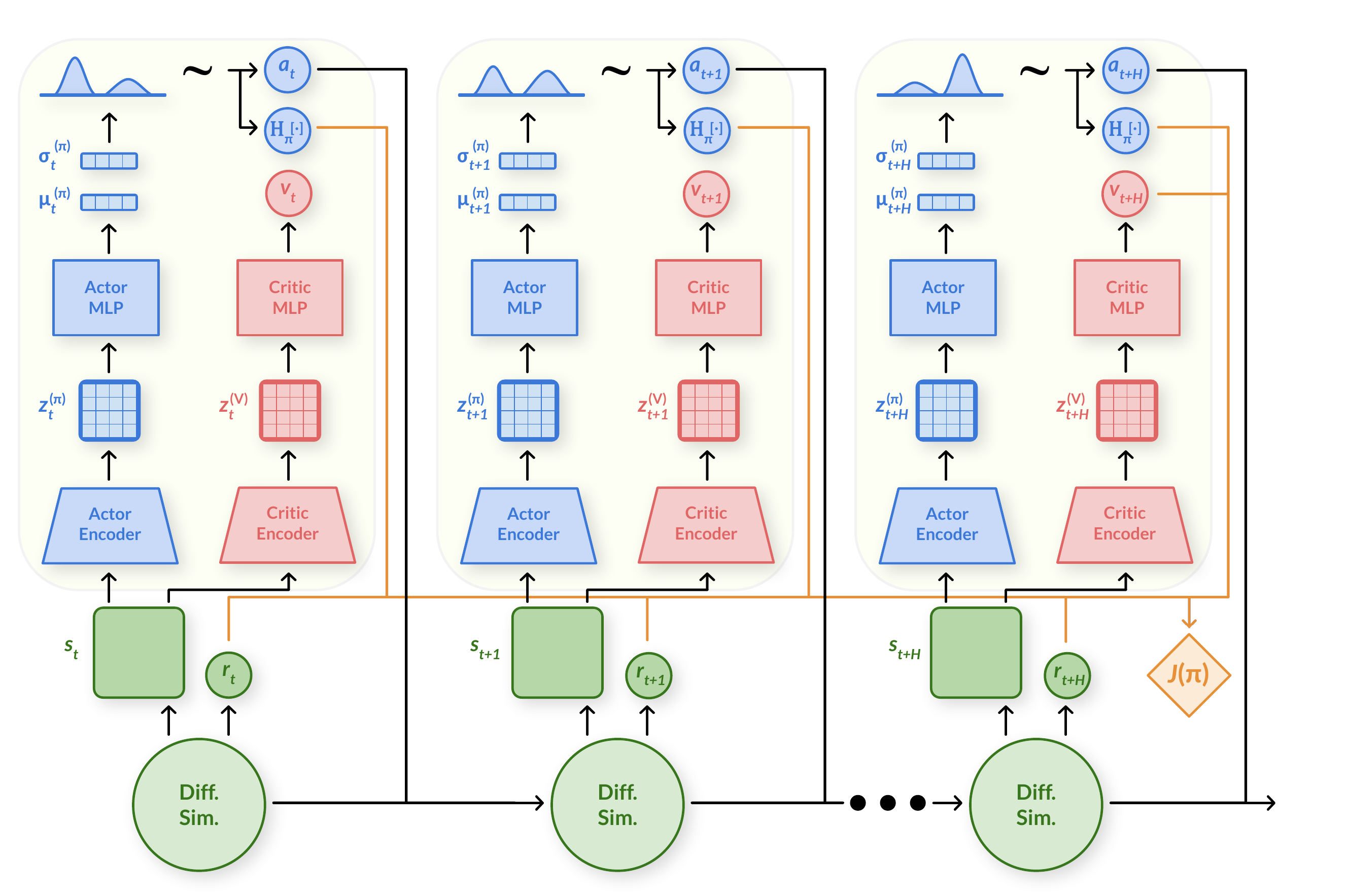}
    \caption{
        \textbf{Diagram of computational graph of \ouralgo.} Note that $J(\pi)$ only updates the actor, although gradients are still propagated through the critic.
    }
    \label{fig:algo_diagram}
\end{figure}

\clearpage
\section{\ouralgo Algorithm Details}

\subsection{Visual encoders in differentiable simulation}
\label{sec:visualencoders}

We use separate visual encoders for the actor $\pi_{\theta}(\bm{a}_t | f_{\phi}(\bm{s}_t))$ and critic $V_{\psi}(f_{\zeta}(\bm{s}_t))$, to enable learning on deformable tasks with high-dimensional (particle-based) visual inputs. To maintain differentiability to compute analytic gradients and reduced memory requirements, we use downsampled particle states of the simulation as point clouds for high-dimensional visual observations. For runtime efficiency, we use a PointNet-like DP3 encoder~\citep{ze20243d} to encode a point cloud observation into a lower-dimensional latent vector. We leave combining differentiable rendering (of RGB or depth image observations) with differentiable simulation, like in~\citep{murthy2021gradsim}, to future work.

\subsection{Pseudocode}
\label{sec:pseudocode}

\begin{algorithm}[H]
\caption{\OURALGO (\ouralgo)}
\label{algo:ours}
\SetKwComment{Comment}{//}{}
\begin{multicols}{2}
Initialize network parameters $\theta,\phi,\psi_i,\zeta_i$\\
$t_0 \gets 0$\\
\Repeat{converged}{
  Create buffer $\mathcal{B}$\\
  \For{$t = t_0 + 1 \ldots H$}{
    $\bm{a}_t \sim \pi_{\theta}(\cdot | f_{\phi}(\bm{s}_t))$\\
    $h_t \gets \mathcal{H}_{\pi}[\bm{a}_t | \bm{s}_t]$\\
    $\hat{h}_t \gets (h_t + \abs{\bar{\mathcal{H}}}) / (2\abs{\bar{\mathcal{H}}})$\\
    $\bm{s}_{t+1}, r_t, d_t \gets$ \texttt{env.step$(\bm{a}_t)$}\\
    $v_{t+1}^{(i)} \gets V_{\psi_i}(f_{\zeta_i}(\bm{s}_{t+1}))$\\
    \If{$d_t$}{
      $t_0 \gets 0$\\
    }
    $\triangleright$ Add data to buffer :\\
    $\mathcal{B} \gets \mathcal{B} \cup \{(\bm{s}_t, \bm{a}_t, r_t, d_t, h_t, \{v_{t+1}^{(i)}\})\}$\\
  }
  $t_0 \gets t_0 + (H + 1)$\\
  \BlankLine
  \BlankLine
  $\triangleright$ Update actor using Eq.~\ref{eq:maxentcostsoftval}, with normalized entropy $\hat{h}_{t}$ and mean values $\frac{1}{C}\sum_{i=1}^C v^{(i)}_t$ :\\
  $(\theta,\phi) \gets (\theta,\phi) - \eta \nabla_{(\theta,\phi)} J_{\mathrm{maxent}}(\pi)$\\
  \BlankLine
  \BlankLine
  $\triangleright$ Detach data from differentiable simulation autograd :\\
  $\mathcal{B} \gets $ \texttt{stopgrad}$(\mathcal{B})$\\
  \BlankLine
  \BlankLine
  $\triangleright$ Update entropy temperature using Eq.~\ref{eq:autoent}, with unnormalized entropy $h_t$ :\\
  $\alpha \gets \alpha - \eta \nabla_{\alpha} [\frac{1}{H}\sum_{t=1}^{H} \alpha (h_t - \bar{\mathcal{H}})]$\\
  \BlankLine
  \BlankLine
  $\triangleright$ Compute TD$(\lambda)$ value targets via Eq.~\ref{eq:tdlambda} using soft returns of Eq.~\ref{eq:soft_bootstrap_ret}, with normalized entropy $\hat{h}_t$ and min values $\min_{i=1\ldots C} v^{(i)}_t$ :\\
  $\tilde{v}_t \gets \ldots$\\
  \BlankLine
  \BlankLine
  \For{$K$ updates}{
    Sample $(\bm{s}_t, \tilde{v}_t) \sim \mathcal{B}$\\
    $\triangleright$ Update critics using Eq.~\ref{eq:shaccriticloss} with clipped soft value targets $\tilde{v}$ :\\
    $(\psi_i,\zeta_i) \gets (\psi_i,\zeta_i) - \eta \nabla_{(\psi_i,\zeta_i)} \mathcal{L}(V)$\\
  }
}
\columnbreak
\textbf{Model components\\}
\begin{tabular}{ll}
    Actor & $\pi_{\theta}(\bm{a}_t | f_{\phi}(\bm{s}_t))$ \\
    Actor encoder & $f_{\phi}(\bm{s}_t)$ \\
    Critic & $V_{\psi_i}(f_{\zeta_i}(\bm{s}_t))$ \\
    Critic encoder & $f_{\zeta_i}(\bm{s}_t)$ \\
    Critic index & $i=1 \ldots C$ \\
\end{tabular}
\BlankLine
\textbf{Hyperparameters\\}
\begin{tabular}{ll}
    Horizon & $H$ \\
    Entropy temperature & $\alpha$ \\
    Target entropy & $\bar{\mathcal{H}}$ \\
    TD trace decay & $\lambda$ \\
    Discount & $\gamma$ \\
    Learning rates & $\eta$ \\
    Num critics & $C$ \\
    Mini-epochs & $K$ \\
\end{tabular}
\end{multicols}
\BlankLine
\BlankLine
\end{algorithm}

\clearpage
\section{Hyperparameters}
\label{sec:hyperparams}

We run all algorithms on consumer workstations with NVIDIA RTX 4090 GPUs. Each run uses a single GPU, on which we run both the GPU-accelerated parallel simulation and optimization loop. We use a recent high-performance implementation of standard model-free RL algorithms which has been validated for parallel simulation~\citep{li2023parallel}. We aim to use common hyperparameter values among algorithms where applicable, such as for discount factor, network architecture, etc.

For TrajOpt, we initialize a single $T$-length trajectory of zero actions. This action is repeated across $N=16$ parallel environments ($N=32$ for $\{$AntRun, HandReorient$\}$). We optimize this trajectory for 50 epochs with a horizon $H$ of 32 steps. We use AdamW as the optimizer, with learning rate of $0.01$, decay rates $(\beta_1, \beta_2) = (0.7, 0.95)$, and gradient norm clipping of $0.5$. For evaluation, we playback this single trajectory across parallel environments, each with different random initial states. 

\begin{table}[H]
\begin{center}
\makebox[\linewidth][c]{%
\setlength{\tabcolsep}{0.2em}
\begin{tabular}{lcccccc}
    \toprule
     & \textit{shared} & PPO & SAC & APG & SHAC & \ouralgo \\
    \midrule
    Num envs $N$ & 64 & & & & & \\
    Batch size & 2048 \\
    Horizon $H$ & 32 \\
    Mini-epochs $K$ & & 5 & 8 & 1 & 16 & 16 \\
    Discount $\gamma$ & $0.99$ \\
    TD/GAE $\lambda$ & - & $0.95$ & & & $0.95$ & $0.95$ \\
    Actor $\eta$ & & \num{5e-4} & \num{5e-4} & \num{2e-3} & \num{2e-3} & \num{2e-3} \\
    Critic $\eta$ & - & \num{5e-4} & \num{5e-4} & & \num{5e-4} & \num{5e-4} \\
    Entropy $\eta$ & - & & \num{5e-3} & & & \num{5e-3} \\
    $\eta$ schedule & - & KL($0.008$) & & linear & linear & linear \\
    Optim type & AdamW & & & Adam & Adam & \\
    Optim $(\beta_1, \beta_2)$ & $(0.9, 0.999)$ & & & $(0.7, 0.95)$ & $(0.7, 0.95)$ & $(0.7, 0.95)$ \\
    Grad clip & $0.5$ & & & $1.0$ & $1.0$ & \\
    Norm type & LayerNorm & & & & & \\
    Act type & SiLU & & & ELU & ELU & \\
    Actor $\sigma(\mathbf{s})$ & yes & & & no & no & \\
    Actor $\log(\sigma)$ & - & $\log [0.1, 1.0]$ & $[-5, 2]$ & & & $[-5, 2]$ \\
    Num critics $C$ & - & & 2 & & & 2 \\
    Critic $\tau$ & - & & $0.995$ & & $0.995$ & \\
    Replay buffer & - & & $10^6$ & & & \\
    Target entropy $\bar{\mathcal{H}}$ & - & & $-\mathrm{dim}(\mathcal{A})/2$ & & & $-\mathrm{dim}(\mathcal{A})/2$ \\
    Init temperature & - & & $1.0$ & & & $1.0$ \\
    \bottomrule
\end{tabular}
}
\end{center}
\caption{\textbf{Shared hyperparameters.} Algorithms use hyperparameter settings in the \textit{shared} column unless otherwise specified in an individual column.}
\label{tab:shared_params}
\vspace{-1.5em}
\end{table}

\begin{table}[H]
\begin{center}
\begin{tabular}{lcccc}
    \toprule
     & Hopper & Ant & Humanoid & SNUHumanoid \\
    \midrule
    Actor MLP & $(128,64,32)$ & $(128,64,32)$ & $(256,128)$ & $(512,256)$  \\
    Critic MLP & $(64,64)$ & $(64,64)$ & $(128,128)$ & $(256,256)$  \\
    \bottomrule
\end{tabular}
\end{center}
\caption{\textbf{DFlex task-specific hyperpararameters.} All algorithms use the same actor and critic network architecture.}
\label{tab:dflex_params}
\vspace{-1.5em}
\end{table}

\begin{table}[H]
\begin{center}
\begin{tabular}{lccc}
    \toprule
     & \textit{shared} & AntRun & HandReorient \\
    \midrule
    Num envs $N$ & 32 & 64 & 64 \\
    Batch size & 1024 & 2048 & 2048 \\
    Actor MLP & $(512,256)$ & $(128,64,32)$ \\
    Critic MLP & $(256,128)$ & $(64,64)$ \\
    \bottomrule
\end{tabular}
\end{center}
\caption{\textbf{\oursim task-specific hyperpararameters.} All algorithms use the same actor and critic network architecture. Algorithms use hyperparameter settings in the \textit{shared} column unless otherwise specified in an individual column.}
\label{tab:rewarped_params}
\vspace{-1.5em}
\end{table}

\clearpage
\section{\oursim Physics}
\label{sec:physics}

We review the simulation techniques used to simulate various rigid and soft bodies in \oursim. Our discussion is based on~\citep{xu2021accelerated, heiden2023disect, murthy2021gradsim, warp2022}, as well as~\citep{hu2019chainqueen, huang2020plasticinelab, ma2023learning} for MPM.

To backpropagate analytic policy gradients, we need to compute simulation gradients. Following reverse order, observe :
\begin{align}
\nabla_{\theta} J(\pi) &= \nabla_{\theta} \sum_{t=0}^{T-1} \gamma^t r_t \\
\frac{\partial r_t}{\partial \theta} &= \frac{\partial r_t}{\partial \bm{a}_t} \frac{\partial \bm{a}_t}{\partial \theta} + \frac{\partial r_t}{\partial \bm{s}_t} \frac{d\bm{s}_t}{d\theta}
\\
\frac{d\bm{s}_{t+1}}{d\theta} &= \frac{\partial f(\bm{s}_t, \bm{a}_t)}{\partial \bm{s}_t} \frac{d \bm{s}_t}{d \theta} + \frac{\partial f(\bm{s}_t, \bm{a}_t)}{\partial \bm{s}_t} \frac{\partial \bm{a}_t}{\partial \theta}
\\
\textrm{with } & \bm{a}_t \sim \pi_{\theta} (\cdot | \bm{s}_t), \bm{s}_{t+1} = f(\bm{s}_t, \bm{a}_t) .
\end{align}
We use the reparameterization trick to compute gradients $\frac{\partial \pi_{\theta} (\bm{s})}{\partial \theta}$ for stochastic policy $\pi_{\theta}$ to obtain $\frac{\partial \bm{a}_t}{\partial \theta}$. Note that the simulation gradient is $\frac{\partial f(\bm{s}_t, \bm{a}_t)}{\partial \bm{s}_t}$, which we compute by the adjoint method through auto-differentiation (AD). 

While we could estimate the simulation gradient with finite differences (FD) through forms of $f'(x) \approx \frac{f(x+ h) - f(x)}{h}$, FD has two major limitations. Consider $f:\mathbb{R}^N \rightarrow \mathbb{R}^M$, where $N$ is the input dimension, $M$ is the output dimension, and $L$ is the cost of evaluating $f$. Then FD has time complexity $O((N + 1) \cdot M \cdot L)$. In comparison, reverse-mode AD has time complexity $O(M\cdot L)$, and also benefits from GPU-acceleration. Furthermore, FD only computes numerical approximations of the gradient with precision depending on the value of $h$, while AD yields analytic gradients. For a simulation platform built on a general differentiable programming framework, computing gradients for \textit{different} simulation dynamics $f$ is straightforward using AD. While alternatives to compute analytic simulation gradients through implicit differentiation have been proposed (see Appendix~\ref{sec:extendedrelatedwork}), they are less amenable to batched gradient computation for parallel simulation.

\subsection{Composite rigid body algorithm (CRBA)}

Consider the following rigid body forward dynamics equation to solve :
\begin{align}
    M\ddot{q} = J^T \mathcal{F}(q, \dot{q}) + c(q, \dot{q}) + \tau(q, \dot{q}, a) ,
\end{align}
where $q, \dot{q}, \ddot{q}$ are joint coordinates, velocities, accelerations, $\mathcal{F}$ are external forces, $c$ includes Coriolis forces, $\tau$ are joint-space actuations, $M$ is the mass matrix, and $J$ is the Jacobian. Featherstone's composite rigid body algorithm (CRBA) is employed to solve for articulation dynamics. After obtaining joint accelerations $\ddot{q}$, a semi-implicit Euler integration step is performed to update the system state $\bm{s}=(q, \dot{q})$. We use the same softened contact model as~\citep{xu2021accelerated}.

\subsection{Finite element method (FEM)}

To simulate dynamics, a finite element model (FEM) is employed based on tetrahedral discretization of the solid's mesh. A stable neo-Hookean constitutive model~\citep{smith2018stable} is used to model elastic deformable solids with per-element actuation :
\begin{align}
    \Psi (q, \tau) = \frac{\mu}{2} (I_C - 3) + \frac{\lambda}{2} (J - \alpha)^2 - \frac{\mu}{2} \log(I_C + 1) ,
\end{align}
where $(\lambda, \mu)$ are the Lam\'e parameters which control each tetrehedral element's resistance to shear and strain, $\alpha$ is a constant, $J=\mathrm{det}(\mathcal{F})$ is the relative volume change, $I_C = \mathrm{tr}(\mathcal{F}^T \mathcal{F})$, and $\mathcal{F}$ is the deformation gradient. Integrating $\Psi$ over each tetrahedral element yields the total elastic potential energy, to then compute $\mathcal{F}_{elastic}$ from the energy gradient, and finally update the system state using a semi-implicit Euler integration step. We use the same approach as~\citep{murthy2021gradsim}.

\subsection{Material point method (MPM)}

The moving least squares material point method (MLS-MPM)~\citep{hu2018moving} can efficiently simulate a variety of complex deformables, such as elastoplastics and liquids. Consider the two equations for conservation of momentum and mass :
\begin{align}
    \rho \ddot{\phi} &= \nabla \cdot P + \rho b \\
    \frac{D \rho}{D t} &= -\rho \nabla \cdot \dot{\phi} ,
\end{align}
where $\dot{\phi}$ is the velocity, $\ddot{\phi}$ is the acceleration, $\rho$ is the density, and $b$ is the body force. For this system to be well-defined, constitutive laws must define $P$, see~\citep{ma2023learning}. We simulate elastoplastics using an elastic constitutive law with von Mises yield criterion :
\begin{align}
    P(F) &= U(2\mu \epsilon + \lambda \mathrm{tr}(\epsilon))U^T \\
    \delta \gamma &= \norm{\hat{\epsilon}} - \frac{2 \sigma_y}{2\mu} \\
    \mathcal{P}(F) &= \begin{cases}
    F & \delta \gamma \leq 0 \\
    U \exp(\epsilon - \delta \gamma \frac{\hat{\epsilon}}{\norm{\hat{\epsilon}}}) V^T & \delta \gamma > 0 ,
    \end{cases}
\end{align}
where $\mathcal{P}(F)$ is a projection back into the elastic region for stress that violates the yield criterion, and $F=U\Sigma V^T$ is the singular value decomposition (SVD). We simulate fluids as weakly compressible, using a plastic constitutive law with the fixed corotated elastic model :
\begin{align}
    P(F) &= \lambda J(J - 1)F^{-T} \\
    \mathcal{P}(F) &= J^{\frac{1}{3}} I ,
\end{align}
where $\mathcal{P}(F)$ is a projection back into the plastic region and $J=\det(F)$.

We use the same softened contact model from PlasticineLab~\citep{huang2020plasticinelab}, with one-way coupling between rigid bodies to MPM particles. For any grid point with signed distance $d$ to the nearest rigid body, we compute a smooth collision strength $s=\min(\exp(-\alpha d), 1)$. The grid point velocity before and after collision projection is linearly interpolated using $s$. Coupling is implemented by the Compatible Particle-In-Cell (CPIC) algorithm, see steps 2--4 of~\citep{hu2018moving}.

\clearpage
\section{\oursim Tasks}
\label{sec:tasks}

We visualize each task in Figure~\ref{fig:oursim_viz}, and high-level task descriptions are provided in Section~\ref{sec:experiments}.

\subsection{Task definitions}

For each task, we define the observation space $\mathcal{O}$, action space $\mathcal{A}$, reward function $R$, termination $d$, episode length $T$, initial state distribution $\rho_0$, and simulation method used for transition function $P$. We denote the goal state distribution $\rho^*$ if the task defines it. We also report important physics hyperparameters, including both shared ones and those specific to the simulation method used.

We use $\hat{i}, \hat{j}, \hat{k}$ for standard unit vectors and $\mathrm{proj}_{w} u = \frac{u \cdot w}{\norm{w}^2} w$ for the projection of $u$ onto $w$. We use $u_x$ to denote the $x$-coordinate of vector $u$.

We use $\mathds{1}$ for the indicator function and $U(b) = U(-b,b)$ for the uniform distribution. We denote an additive offset ${}^{+}\delta$ or multiplicative scaling ${}^{\times}\delta$, from some prior value.

For notation, we use $z$-axis up. Let $q, \dot{q}, \ddot{q}$ be joint coordinates, velocities, accelerations. Let $(p, \theta)$ be world positions and orientations, $(v, \omega)$ be linear and angular velocities, $(a, \alpha)$ be linear and angular accelerations, derived for root links such as the center of mass (CoM). 

All physical quantities are reported in SI units. We use $h$ for the frame time, and each frame is computed using $S$ substeps, so the physics time step is $\Delta t = \frac{h}{S}$. The gravitational acceleration is $g$.

\textbf{FEM.} We use $\bm{x}'$ to denote a subsampled set of FEM particle positions $\bm{x}$, where $\overline{\bm{x}'}$ is the CoM of the subsampled particles (i.e. average particle position for uniform density). Similarly for velocities $\bm{v}$. We also report the following quantities: number of particles $N_{\bm{x}}$, number of tetrahedron $N_{tet}$, particles' density $\rho$, Lam\'e parameters $(\lambda, \mu)$ and damping stiffness $k_{damp}$.

\textbf{MPM.} We use $\bm{x}'$ to denote a subsampled set of MPM particle positions $\bm{x}$, where $\overline{\bm{x}'}$ is the CoM of the subsampled particles (i.e. average particle position for uniform density). We also report the following quantities: friction coefficient for rigid bodies $\mu_b$, grid size $N_g$, number of particles $N_{\bm{x}}$, particles' density $\rho$, Young's modulus $E$, Poisson's ratio $\nu$, yield stress $\sigma_y$.

\clearpage
\subsubsection{AntRun}

\begin{table}[h]
\begin{center}
{\renewcommand{\arraystretch}{1.25}
\begin{tabular}{l|c}
    \toprule
    \midrule
    $\mathcal{O}$ & $\mathbb{R}^{37}$ : $[p_z, \theta, v, \omega, q, \dot{q}, u_{up}, u_{heading}, \bm{a}_{t-1}]$ \\
    $\mathcal{A}$ & $\mathbb{R}^8$ : absolute joint torques $\tau$ \\
    $R$ & $v_x + (0.1 R_{up} + R_{heading} + R_{height})$ \\
    $d$ & $\mathds{1}\{p_z < h_{term}\}$ \\
    $T$ & 1000 \\
    \midrule
    $\rho_0$ & ${}^{+}\delta p\sim U(0.1), {}^{+}\delta \theta \sim U(\frac{\pi}{24})$ \\
    & ${}^{+}\delta q \sim U(0.2)$ \\ 
    & $\dot{q} \sim U(0.25)$ \\
    \midrule
    $P$ & CRBA \\
    $h$ & $1/60$ \\
    $S$ & 16 \\
    \bottomrule
\end{tabular}
}
\end{center}
\caption{\textbf{AntRun task definition}. Based on Ant from~\citep{xu2021accelerated}.}
\end{table}

The derived quantities correspond to the ant's CoM. We use termination height $h_{term}=0.27$. We denote $u_{up} = \mathrm{proj}_{\hat{k}\:} p$ and $u_{heading} = \mathrm{proj}_{\hat{i}\:} p$. The reward function maximizes the forward velocity $v_x$, with auxiliary reward terms $R_{up} = u_{up}$ to encourage vertical stability, $R_{heading} = u_{heading}$ to encourage running straight, and $R_{height} = p_z - h_{term}$ to discourage falling. Initial joint angles, joint velocities, and CoM transform of the ant are randomized.

\clearpage
\subsubsection{HandReorient}

\begin{table}[h]
\begin{center}
{\renewcommand{\arraystretch}{1.25}
\begin{tabular}{l|c}
    \toprule
    \midrule
    $\mathcal{O}$ & $\mathbb{R}^{72}$ : ``full'', see~\citep{makoviychuk2021isaac} \\
    $\mathcal{A}$ & $\mathbb{R}^{16}$ : absolute joint torques $\tau$ \\
    $R$ & see~\citep{makoviychuk2021isaac} \\
    $d$ & see~\citep{makoviychuk2021isaac} \\
    $T$ & 600 \\
    \midrule
    $\rho_0$ & see~\citep{makoviychuk2021isaac} \\
    \midrule
    $P$ & CRBA \\
    $h$ & $1/120$ \\
    $S$ & 32 \\
    \bottomrule
\end{tabular}
}
\end{center}
\caption{\textbf{HandReorient task definition}. Based on AllegroHand from~\citep{makoviychuk2021isaac}.}
\end{table}

See~\citep{makoviychuk2021isaac}. Initial joint angles of the hand, CoM transform of the cube, and target orientations for the cube are randomized.

\clearpage
\subsubsection{RollingFlat}

\begin{table}[h]
\begin{center}
{\renewcommand{\arraystretch}{1.25}
\begin{tabular}{l|c}
    \toprule
    \midrule
    $\mathcal{O}$ & $[\mathbb{R}^{250 \times 3}, \mathbb{R}^3, \mathbb{R}^{3}]$ : $[\bm{x}', \overline{\bm{x}'}, (p_x, p_z, \theta_z)]$ \\
    $\mathcal{A}$ & $\mathbb{R}^3$ : relative positions $p_x, p_z$ and orientation $\theta_z$ \\
    $R$ & $R_{d} + R_{flat}$ \\
    $d$ & - \\
    $T$ & 300 \\
    \midrule
    $\rho_0$ & ${}^{\times}\delta \bm{x} \sim U(0.9, 1.1)$ \\
    & ${}^{+}\delta \bm{\overline{x}}_{x,y} \sim U(0.1), {}^{+}\delta \bm{\overline{x}}_z \sim U(0, 0.05)$ \\
    \midrule
    $P$ & MPM \\
    $h$ & $S\:\cdot\:$\num{5e-5} \\
    $S$ & 40 \\
    \midrule
    $\mu_b$ & $0.9$ \\
    $N_{g}$ & 48$^3$ \\
    $N_{\bm{x}}$ & 2592 \\
    $\rho$ & $1.0$ \\
    $E$ & $5000.$ \\
    $\nu$ & $0.2$ \\
    $\sigma_y$ & $50.$ \\
    \bottomrule
\end{tabular}
}
\end{center}
\caption{\textbf{RollingFlat task definition}. Based on RollingPin from~\citep{huang2020plasticinelab}.}
\end{table}

The derived quantities correspond to the rolling pin's CoM. We use target flatten height $h_{flat}=0.125$. For the reward function, the first term $R_{d}$ minimizes the difference between the particles' CoM and $h_{flat}$, while the second term maximizes the smoothness of the particles $R_{flat}$ :
\begin{align*}
    d &= \frac{\overline{\bm{x}'}_z}{h_{flat}} \\
    R_d &= \Big(\frac{1}{1 + d}\Big)^2 [\mathds{1}\{d > 0.33\}\cdot 1 + \mathds{1}\{d \leq 0.33\} \cdot 2] \\
    R_{flat} &= -\Var[\bm{x}'_z]
\end{align*}
Initial volume and CoM transform of the particles are randomized.

\clearpage
\subsubsection{SoftJumper}

\begin{table}[h]
\begin{center}
{\renewcommand{\arraystretch}{1.25}
\begin{tabular}{l|c}
    \toprule
    \midrule
    $\mathcal{O}$ & $[\mathbb{R}^{204 \times 3}, \mathbb{R}^3, \mathbb{R}^3, \mathbb{R}^{222}]$ : $[\bm{x}', \overline{\bm{x}'}, \overline{\bm{v}'}, \bm{a}_{t-1}]$ \\
    $\mathcal{A}$ & $\mathbb{R}^{222}$ : absolute tetrahedral volumetric activations (subsampled) \\
    $R$ & $\overline{\bm{v}'}_x + (3.0R_{up} - 0.0001 \sum \norm{\bm{a}}_2^2)$ \\
    $d$ & - \\
    $T$ & 300 \\
    \midrule
    $\rho_0$ & ${}^{\times}\delta \bm{x} \sim U(0.9, 1.1)$ \\
    & ${}^{+}\delta \bm{\overline{x}}_{x,y} \sim U(0.8), {}^{+}\delta \bm{\overline{x}}_z \sim U(0, 0.4)$ \\
    \midrule
    $P$ & FEM \\
    $h$ & $1/60$ \\
    $S$ & 80 \\
    \midrule
    $N_{\bm{x}}$ & 204 \\
    $N_{tet}$ & 444 \\
    $\rho$ & $1.0$ \\
    $\lambda$ & $1000.$ \\
    $\mu$ & $1000.$ \\
    $k_{damp}$ & $1.0$ \\
    \bottomrule
\end{tabular}
}
\end{center}
\caption{\textbf{SoftJumper task definition}. Inspired by~\citep{murthy2021gradsim, hu2020difftaichi}.}
\end{table}

Denote the default initial height $h_0 = \overline{\bm{x}}_z = 0.2$. The reward function maximizes the forward velocity $\overline{\bm{v}'}_x$ of the quadruped's CoM, with two auxiliary reward terms $R_{up} = \overline{\bm{x}'}_z - h_0$ to encourage jumping and the other to encourage energy-efficient policies by minimizing action norms. For the action space, we downsample by a factor of $2$ and upsample actions to obtain the full set of activations. Initial volume and CoM transform of the quadruped are randomized.

\clearpage
\subsubsection{HandFlip}
\label{sec:taskdef_handflip}

\begin{table}[h]
\begin{center}
{\renewcommand{\arraystretch}{1.25}
\begin{tabular}{l|c}
    \toprule
    \midrule
    $\mathcal{O}$ & $[\mathbb{R}^{250 \times 3}, \mathbb{R}^3, \mathbb{R}^{24}]$ : $[\bm{x}', \overline{\bm{x}'}, q]$ \\
    $\mathcal{A}$ & $\mathbb{R}^{24}$ : relative joint angles \\
    $R$ & $NI\{ R_{flip} \}$ \\
    $d$ & - \\
    $T$ & 300 \\
    \midrule
    $\rho_0$ & ${}^{\times}\delta \bm{x} \sim U(0.95, 1.05)$ \\
    & ${}^{+}\delta \bm{\overline{x}}_{x,y} \sim U(0.1), {}^{+}\delta \bm{\overline{x}}_z \sim U(0, 0.1)$ \\
    \midrule
    $P$ & MPM \\
    $h$ & $S\:\cdot\:$\num{5e-5} \\
    $S$ & 40 \\
    \midrule
    $\mu_b$ & $0.5$ \\
    $N_{g}$ & 48$^3$ \\
    $N_{\bm{x}}$ & 2500 \\
    $\rho$ & $1.0$ \\
    $E$ & $4000.$ \\
    $\nu$ & $0.2$ \\
    $\sigma_y$ & $130.$ \\
    \bottomrule
\end{tabular}
}
\end{center}
\caption{\textbf{HandFlip task definition}. Based on Flip from~\citep{li2023dexdeform}.}
\end{table}

We define $NI\{r\} = \mathrm{clamp}(\frac{r_0 - r}{r_0}, -1.0, 1.0)$ as the normalized improvement of $r$ from initial $r_0$. The reward function maximizes the $NI$ of $R_{flip}$, where the term $R_{flip} = \overline{\norm{[\bm{x}']_{h_l} - [\bm{x}']_{h_u}}}_2$ minimizes the average Euclidean distance between two mirrored halves $h_{l}, h_{u}$ of the particles. Initial volume and CoM transform of the particles are randomized.

\clearpage
\subsubsection{FluidMove}

\begin{table}[h]
\begin{center}
{\renewcommand{\arraystretch}{1.25}
\begin{tabular}{l|c}
    \toprule
    \midrule
    $\mathcal{O}$ & $[\mathbb{R}^{250 \times 3}, \mathbb{R}^3, \mathbb{R}^{3}, \mathbb{R}^{3}]$ : $[\bm{x}', \overline{\bm{x}'}, p, p^*]$ \\
    $\mathcal{A}$ & $\mathbb{R}^3$ : relative positions $p$ \\
    $R$ & $R_d + R_{spill}$ \\
    $d$ & - \\
    $T$ & 300 \\
    \midrule
    $\rho_0$ & - \\
    $\rho^*$ & $p^* \leftarrow [{}^{+}\delta p_{x,y} \sim U(0.2), {}^{+}\delta p_z \sim U(0.0, 0.3)]$ \\
    \midrule
    $P$ & MPM \\
    $h$ & $S\:\cdot\:$\num{5e-4} \\
    $S$ & 40 \\
    \midrule
    $\mu_b$ & $0.5$ \\
    $N_{g}$ & 48$^3$ \\
    $N_{\bm{x}}$ & 2880 \\
    $\rho$ & $1e3$ \\
    $E$ & $1e5$ \\
    $\nu$ & $0.3$ \\
    \bottomrule
\end{tabular}
}
\end{center}
\caption{\textbf{FluidMove task definition}. Based on TransportWater from~\citep{lin2021softgym}.}
\end{table}

For the reward function, the first term $R_d$ minimizes the distance to the target position $p^{*}$, while the second term penalizes particles which are spilled outside the container :
\begin{align*}
    d &= \norm{p - p^*}^2_2 \\
    R_d &= \Big(\frac{1}{1 + d}\Big)^2 [\mathds{1}\{d > 0.02\}\cdot 1 + \mathds{1}\{d \leq 0.02\} \cdot 2] \\
    R_{spill} &= \textrm{see~\citep{lin2021softgym}}
\end{align*}

Target position for the container is randomized.

\clearpage
\section{Extended Experimental Results}

\subsection{Example of SHAC getting stuck in local optima}
\label{sec:shacsuboptimal}

\begin{figure}[H]
    \centering
    \vspace{-1.25em}
    \includegraphics[width=0.32\columnwidth]{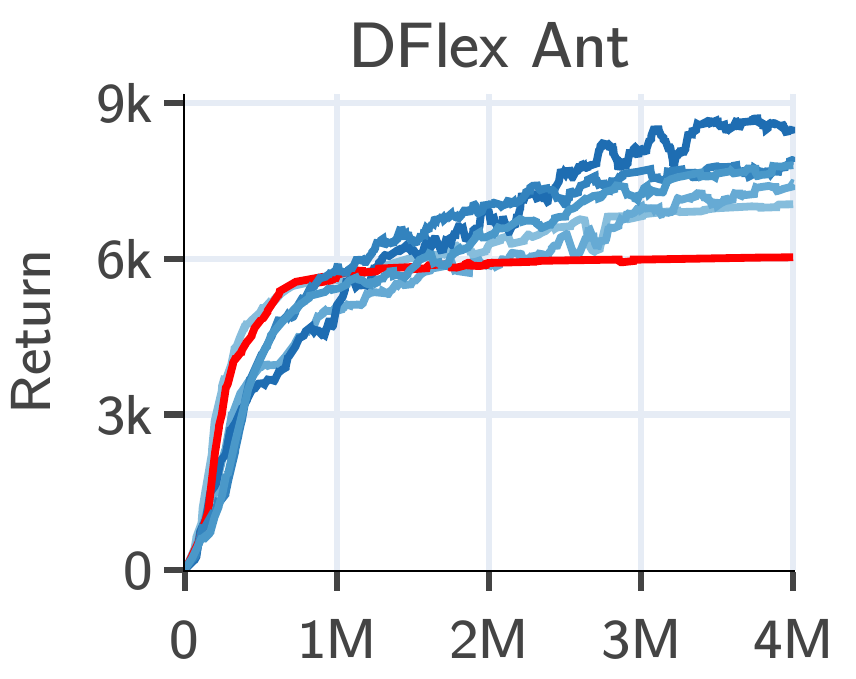}
    \caption{
        \textbf{Example of SHAC getting stuck in local minima.}
        Episode return as a function of environment steps in DFlex Ant ($\mathcal{A}\subset\mathbb{R}^{8}$). One run (colored in red) quickly plateaus after 1M steps and does not improve. 6 random seeds.
    }
    \label{fig:shac_individualruns_dflexant4m}
\end{figure}

We reproduce the original DFlex Ant results from SHAC~\citep{xu2021accelerated}, and in Figure~\ref{fig:shac_individualruns_dflexant4m} we visualize individual runs for insight~\citep{patterson2023empirical}. From this, we observe that one of the runs quickly plateaus to a suboptimal policy after 1M steps and does not improve.

\subsection{Results on DFlex locomotion}
\label{sec:dflexlocomotion}

\begin{figure}[H]
    \centering
    \includegraphics[width=\columnwidth]{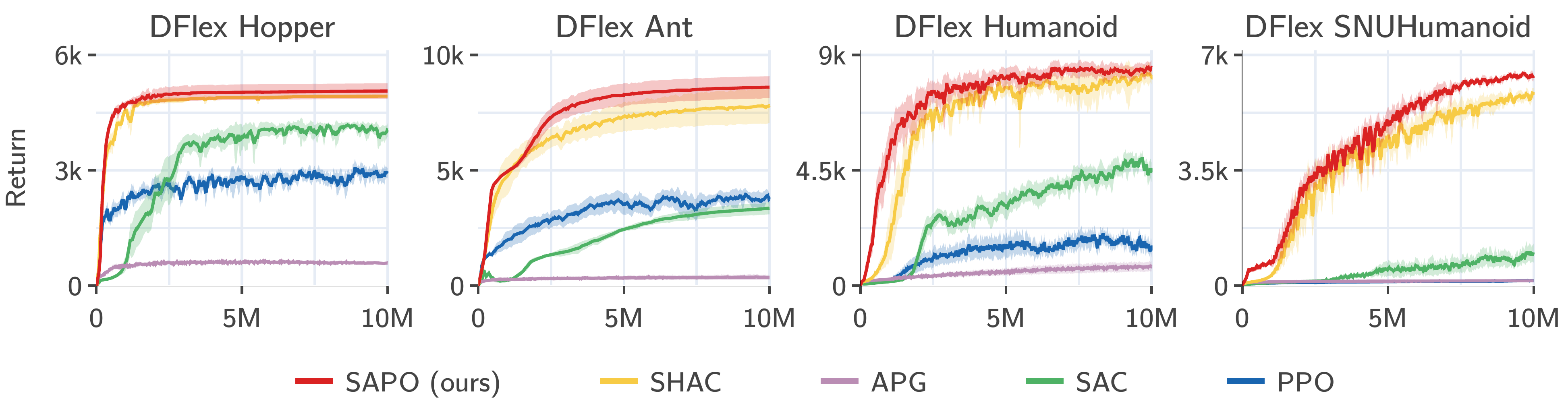}
    \caption{
        \textbf{DFlex locomotion training curves.} 
        Episode return as a function of environment steps in DFlex Hopper ($\mathcal{A}\subset\mathbb{R}^{3}$), Ant ($\mathcal{A}\subset\mathbb{R}^{8}$), Humanoid ($\mathcal{A}\subset\mathbb{R}^{21}$), and SNUHumanoid ($\mathcal{A}\subset\mathbb{R}^{152}$) locomotion tasks. Mean and 95\% CIs over 10 random seeds.
    }
    \label{fig:dflexlocomotion}
\end{figure}

\begin{table}[H]
\begin{center}
\setlength{\tabcolsep}{0.5em}
\begin{tabular}{lcccc}
    \toprule
     & \textbf{Hopper} & \textbf{Ant} & \textbf{Humanoid} & \textbf{SNUHumanoid}\\
    \midrule
    PPO & 3155 $\pm$ 30 & 3883 $\pm$ 60 & 414 $\pm$ 45 & 135 $\pm$ 3 \\
    SAC & 3833 $\pm$ 50 & 3366 $\pm$ 25 & 4628 $\pm$ 120 & 846 $\pm$ 44 \\
    APG & 590 $\pm$ 3 & 368 $\pm$ 11 & 783 $\pm$ 16 & 149 $\pm$ 1 \\
    SHAC & 4939 $\pm$ 3 & 7779 $\pm$ 70 & 8256 $\pm$ 74 & 5755 $\pm$ 67 \\
    \rowcolor{rowblue!10}
    \ouralgo (ours) & 5060 $\pm$ 18 & 8610 $\pm$ 40 & 8469 $\pm$ 58 & 6427 $\pm$ 53 \\
    \bottomrule
\end{tabular}
\end{center}
\caption{\textbf{DFlex locomotion tabular results.} Evaluation episode returns for final policies after training. Mean and 95\% CIs over 10 random seeds with 128 episodes per seed.}
\label{tab:dflexlocomotion}
\end{table}

\subsection{\ouralgo ablations on DFlex locomotion}
\label{sec:ablation_dflex}

\begin{figure}[H]
    \centering
    \includegraphics[width=\columnwidth]{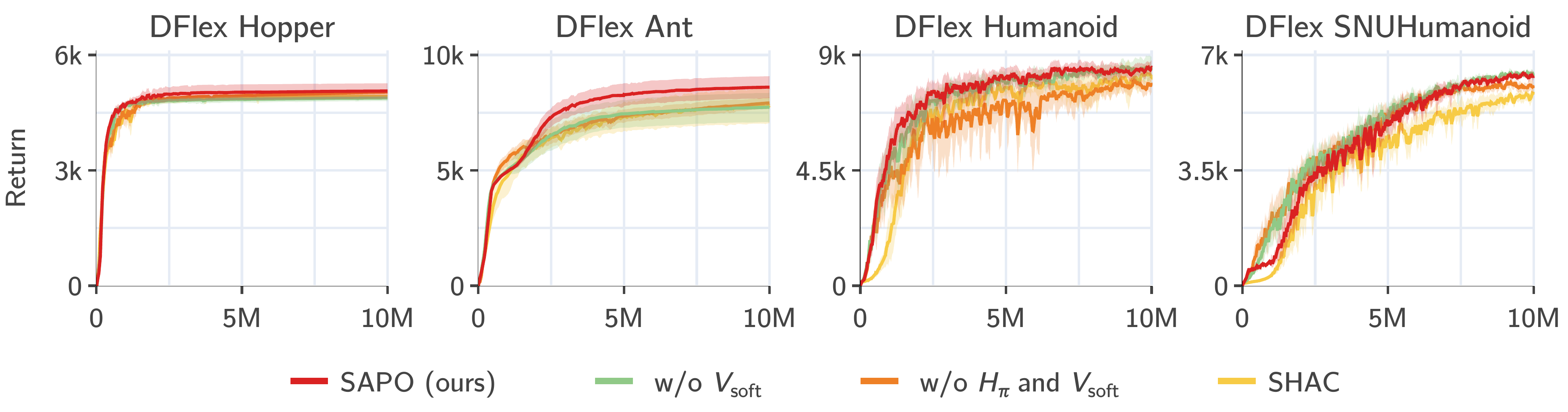}
    \caption{
        \textbf{\ouralgo ablations -- DFlex locomotion training curves.} 
        Episode return as a function of environment steps in DFlex Hopper ($\mathcal{A}\subset\mathbb{R}^{3}$), Ant ($\mathcal{A}\subset\mathbb{R}^{8}$), Humanoid ($\mathcal{A}\subset\mathbb{R}^{21}$), and SNUHumanoid ($\mathcal{A}\subset\mathbb{R}^{152}$) locomotion tasks. Mean and 95\% CIs over 10 random seeds.
    }
    \label{fig:ablation_dflex}
\end{figure}

\begin{table}[H]
\begin{center}
\setlength{\tabcolsep}{0.5em}
\begin{tabular}{lccccc}
    \toprule
     & \textbf{Hopper} & \textbf{Ant} & \textbf{Humanoid} & \textbf{SNUHumanoid} & (\textbf{avg} $\Delta$\%) \\
    \midrule
    \ouralgo (ours) & 5060 $\pm$ 18 & 8610 $\pm$ 42 & 8469 $\pm$ 59 & 6427 $\pm$ 52 & 6.8\% \\
    \:\:w/o $V_{\mathrm{soft}}$ & 4882 $\pm$ 7 & 7729 $\pm$ 52 & 8389 $\pm$ 76 & 6392 $\pm$ 54 & 2.7\% \\
    \:\:w/o $\mathcal{H}_{\pi}$ and $V_{\mathrm{soft}}$ & 5036 $\pm$ 2 & 7897 $\pm$ 30 & 7731 $\pm$ 91 & 6032 $\pm$ 58 & 0.5\% \\
    SHAC & 4939 $\pm$ 3 & 7779 $\pm$ 69 & 8256 $\pm$ 76 & 5755 $\pm$ 66 & -- \\
    \bottomrule
\end{tabular}
\end{center}
\caption{\textbf{\ouralgo ablations -- DFlex locomotion tabular results.} Evaluation episode returns for final policies after training. Mean and 95\% CIs over 10 random seeds with 128 episodes per seed.}
\label{tab:ablation_dflex}
\end{table}

\subsection{Additional \ouralgo ablations for design choices \{III, IV, V\}}
\label{sec:ablation_designchoices}

\begin{figure}[H]
    \begin{minipage}{0.51\columnwidth}
        \centering
        \includegraphics[width=1.0\textwidth,trim={0 0 3cm 0},clip]{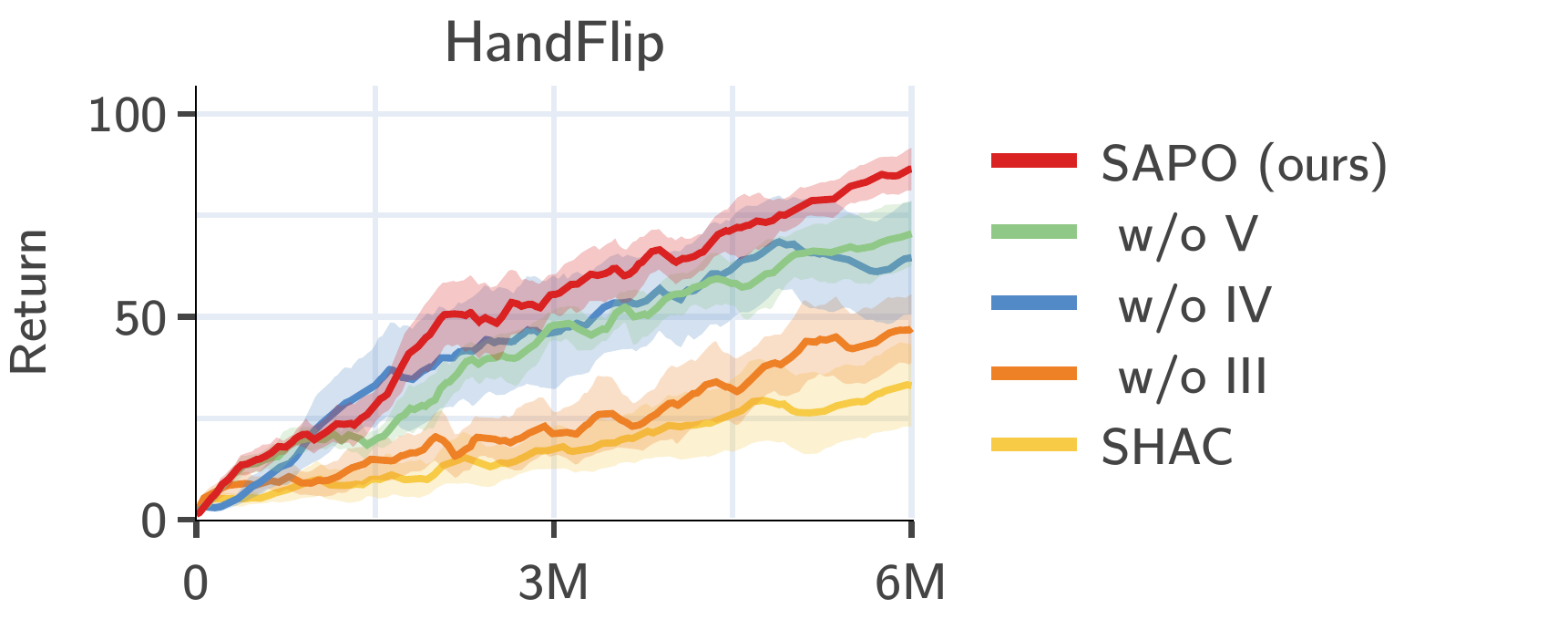}
        \captionof{figure}{
            \textbf{\ouralgo ablations -- (ext.) HandFlip training curves.} Episode return as a function of environment steps. Smoothed using EWMA with $\alpha=0.99$. Mean and 95\% CIs over 10 random seeds.
        }
        \label{fig:ablation_designchoices}
    \end{minipage}%
    \hspace{0.5em}
    \begin{minipage}{0.47\columnwidth}
        \centering
        \begin{table}[H]
        \begin{center}
        \setlength{\tabcolsep}{0.25em}
        \vspace{-1.0em}
        \begin{tabular}{lcc}
            \toprule
             & \textbf{HandFlip} & ($\Delta$\%) \\
            \midrule
            \ouralgo (ours) & 90 $\pm$ 2 & +172.7\% \\
            \:\:w/o V & 71 $\pm$ 3 & +115.2\% \\
            \:\:w/o IV & 64 $\pm$ 3 & \:\:+93.9\% \\
            \:\:w/o III & 49 $\pm$ 3 & \:\:+48.5\% \\
            \midrule
            SHAC & 33 $\pm$ 3 & -- \\
            \bottomrule
        \end{tabular}
        \end{center}
        \captionof{table}{\textbf{\ouralgo ablations -- (ext.) HandFlip tabular results.} Evaluation episode returns for final policies after training. Mean and 95\% CIs over 10 random seeds with 64 episodes per seed.}
        \label{tab:ablation_designchoices}
        \end{table}
    \end{minipage}
\end{figure}

See Section~\ref{sec:designchoices} for descriptions of these design choices.

\clearpage
\subsection{Runtime and scalability of \oursim}

We report all timings on a consumer workstation with an AMD Threadripper 5955WX CPU, NVIDIA RTX 4090 GPU, and 128GB DDR4 3200MHz RAM. We test the scalability of Rewarped's parallel MPM implementation over an increasing number of environments, on the HandFlip task. See Appendix~\ref{sec:taskdef_handflip} for details on environment settings.

We run 5 full-length trajectories, and report the averaged runtime of forward simulation.
As a reference, the non-parallel MPM implementation from~\citep{li2023dexdeform} runs with a single environment at $70$ FPS using $3.51$ GB of GPU memory (when $N_x=2500, N_g=48^3$).

\begin{figure}[H]
\centering
\begin{tikzpicture}
    \definecolor{myblue}{RGB}{25,101,176}
    \definecolor{mygreen}{RGB}{78,178,101}
    \definecolor{myblack}{RGB}{68,68,68}
    \definecolor{gridgray}{RGB}{230,236,245}
    \pgfplotsset{every axis plot/.append style={color=myblack}}

    \begin{axis}[
        title={\large HandFlip},
        xlabel={\# of Envs.},
        ylabel={Total FPS},
        y label style={color=myblue},
        yticklabel style={color=myblue},
        tick align=outside,
        tick style={color=myblack,line width=1.5pt},
        ytick style={color=myblue},
        grid=both,
        grid style={color=gridgray,line width=1.5pt},
        xtick={1,8,16,32,64,128},
        xticklabels={1,8,16,32,64,128},
        ymin=0, ymax=1800,
        axis y line*=left,
        axis x line*=bottom,
        title style={font=\myfont},
        tick label style={/pgf/number format/assume math mode=true, font=\myfont},
        every axis label={font=\myfont},
        label style={font=\myfont},
        legend style={font=\myfont},
    ]
    \addplot[
        color=myblue,
        mark=*,
        line width=1.5pt
    ]
    coordinates {
        (1,210)
        (2,373)
        (4,617)
        (8,978)
        (16,1237)
        (32,1394)
        (64,1506)
        (128,1569)
    };
    \end{axis}

  \begin{axis}[
    ylabel={GPU Memory (GB)},
    y label style={color=mygreen},
    yticklabel style={color=mygreen},
    tick align=outside,
    tick style={color=myblack,line width=1.5pt},
    ytick style={color=mygreen},
    ymin=0, ymax=24,
    axis y line*=right,
    axis x line=none,
    title style={font=\myfont},
    tick label style={/pgf/number format/assume math mode=true, font=\myfont},
    every axis label={font=\myfont},
    label style={font=\myfont},
    legend style={font=\myfont},
  ]
    \addplot[
      color=mygreen,
      mark=square*,
        line width=1.5pt
    ] coordinates {
      (1,1.09)
      (2,1.28)
      (4,1.58)
      (8,2.25)
      (16,3.59)
      (32,6.16)
      (64,11.40)
      (128,21.86)
    };
  \end{axis}
\end{tikzpicture}
\caption{\textbf{Rewarped runtimes -- HandFlip curves.} Total FPS (blue) and GPU memory (green) are shown. Total FPS averaged over $5N$ full-length trajectories with random actions.
}
\label{fig:sim_runtime}
\end{figure}
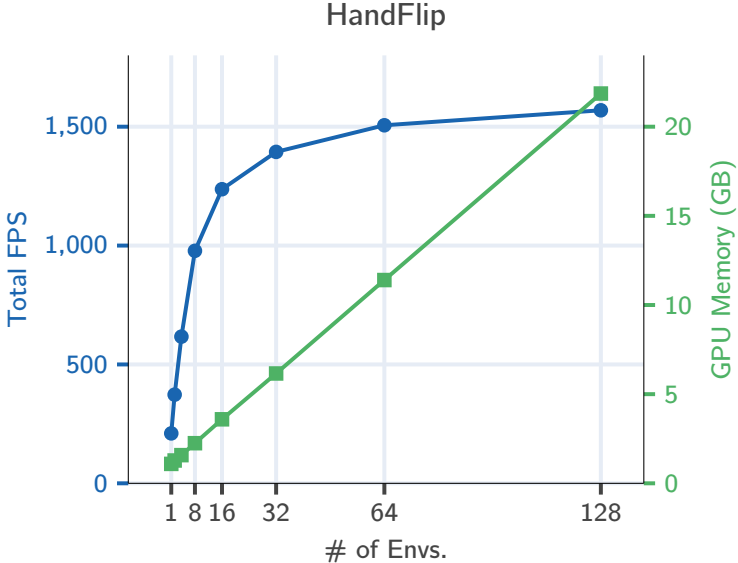

With one environment, Rewarped's MPM implementation is $3\times$ faster and uses $2/3$ less memory compared to DexDeform's. Using $32$ environments, Rewarped achieves a $20\times$ total speedup compared to DexDeform for simulating the HandFlip task. Furthermore, the GPU memory used \textit{per env} decreases because overhead is amortized across multiple environments that are simulated in same underlying physics scene. In Figure~\ref{fig:sim_runtime}, we observe that the total FPS drastically increases until plateauing beyond $N=32$ environments. This occurs because GPU utilization is maxed out; more GPUs or a GPU with more CUDA cores would be needed to continue scaling.

\begin{table}[H]
\begin{center}
\setlength{\tabcolsep}{0.25em}
\begin{tabular}{cccc}
    \toprule
    \# of Envs. & Total FPS & Relative Speedup & GPU Memory (GB) \\
    \midrule
    1 & $210$ & $1.00$ & $1.09$ \\
    2 & $373$ & $1.78$ & $1.28$ \\
    4 & $617$ & $2.94$ & $1.58$ \\
    8 & $978$ & $4.66$ & $2.25$ \\
    16 & $1,237$ & $5.89$ & $3.59$ \\
    32 & $1,394$ & $6.64$ & $6.16$ \\
    64 & $1,506$ & $7.17$ & $11.40$ \\
    128 & $1,569$ & $7.47$ & $21.86$ \\
    \bottomrule
\end{tabular}
\end{center}
\captionof{table}{\textbf{Rewarped runtimes -- HandFlip tabular results.} Total FPS averaged over $5N$ full-length trajectories with random actions.
}
\label{tab:sim_runtime}
\end{table}

\begin{figure}[H]
\centering
\begin{tikzpicture}
    \definecolor{myblue}{RGB}{25,101,176}
    \definecolor{mygreen}{RGB}{78,178,101}
    \definecolor{myblack}{RGB}{68,68,68}
    \definecolor{gridgray}{RGB}{230,236,245}
    \pgfplotsset{every axis plot/.append style={color=myblack}}

    \begin{axis}[
        title={\large HandFlip},
        xlabel={\# of Envs.},
        ylabel={Total FPS},
        y label style={color=myblue},
        yticklabel style={color=myblue},
        tick align=outside,
        tick style={color=myblack,line width=1.5pt},
        ytick style={color=myblue},
        grid=both,
        grid style={color=gridgray,line width=1.5pt},
        xtick={1,8,16,32,64,128},
        xticklabels={1,8,16,32,64,128},
        ymin=0, ymax=200,
        axis y line*=left,
        axis x line*=bottom,
        title style={font=\myfont},
        tick label style={/pgf/number format/assume math mode=true, font=\myfont},
        every axis label={font=\myfont},
        label style={font=\myfont},
        legend style={font=\myfont},
    ]
    \addplot[
        color=myblue,
        mark=*,
        line width=1.5pt
    ]
    coordinates {
        (1,85)
        (2,118)
        (4,150)
        (8,169)
        (16,172)
        (32,179)
    };
    \end{axis}

  \begin{axis}[
    ylabel={GPU Memory (GB)},
    y label style={color=mygreen},
    yticklabel style={color=mygreen},
    tick align=outside,
    tick style={color=myblack,line width=1.5pt},
    ytick style={color=mygreen},
    ymin=0, ymax=24,
    axis y line*=right,
    axis x line=none,
    title style={font=\myfont},
    tick label style={/pgf/number format/assume math mode=true, font=\myfont},
    every axis label={font=\myfont},
    label style={font=\myfont},
    legend style={font=\myfont},
  ]
    \addplot[
      color=mygreen,
      mark=square*,
        line width=1.5pt
    ] coordinates {
      (1,1.50)
      (2,2.20)
      (4,3.12)
      (8,5.37)
      (16,10.22)
      (32,18.71)
    };
  \end{axis}
\end{tikzpicture}
\caption{\textbf{Rewarped runtimes -- HandFlip $(N_x=30000)$ curves.} Total FPS (blue) and GPU memory (green) are shown. Total FPS averaged over $5N$ full-length trajectories with random actions.
}
\label{fig:sim_runtime2}
\end{figure}

\begin{table}[H]
\begin{center}
\setlength{\tabcolsep}{0.25em}
\begin{tabular}{cccc}
    \toprule
    \# of Envs. & Total FPS & Relative Speedup & GPU Memory (GB) \\
    \midrule
    1 & $85$ & $1.00$ & $1.50$ \\
    2 & $118$ & $1.39$ & $2.20$ \\
    4 & $150$ & $1.76$ &$3.12$ \\
    8 & $169$ & $1.99$ &$5.37$ \\
    16 & $172$ & $2.02$ & $10.22$ \\
    32 & $179$ & $2.11$ & $18.71$ \\
    \bottomrule
\end{tabular}
\end{center}
\captionof{table}{\textbf{\oursim runtimes -- HandFlip $(N_x=30000)$ tabular results.} Total FPS averaged over $5N$ full-length trajectories with random actions.
}
\label{tab:sim_runtime2}
\end{table}

In Table~\ref{tab:algo_runtime}, we report training runtimes for algorithms on the HandFlip task, averaged over 10 random seeds. We also note that runtimes may be further optimized using PyTorch compile and CUDA graphs, but we did not implement this.

\begin{table}[H]
\begin{center}
\setlength{\tabcolsep}{0.25em}
\begin{tabular}{lc}
    \toprule
    Algorithm & Runtime (hours) \\
    \midrule
    PPO & $1.46$ \\
    SAC & $1.67$ \\
    APG & $7.43$ \\
    SHAC & $7.66$ \\
    \ouralgo (ours) & $7.77$ \\
    \bottomrule
\end{tabular}
\end{center}
\captionof{table}{\textbf{Training runtimes -- HandFlip task.} Algorithm training runtimes over 6M environment steps. Mean over 10 random seeds.
}
\label{tab:algo_runtime}
\end{table}

\clearpage
\subsection{Visualizations of trajectories in \oursim tasks}
\label{sec:trajviz}

\begin{figure}[ht]
    \centering
    \captionsetup[subfigure]{labelformat=empty,skip=2pt}
    \renewcommand\thesubfigure{}
    \captionsetup[subfigure]{position=above}
    \setlength{\tabcolsep}{0.1em}
    \def\arraystretch{2.5}
    \vspace{-1.0em}
    \begin{tabular}{ccc}
    \subcaptionbox{\myfont\normalsize{}}{\includegraphics[width=0.29\columnwidth]{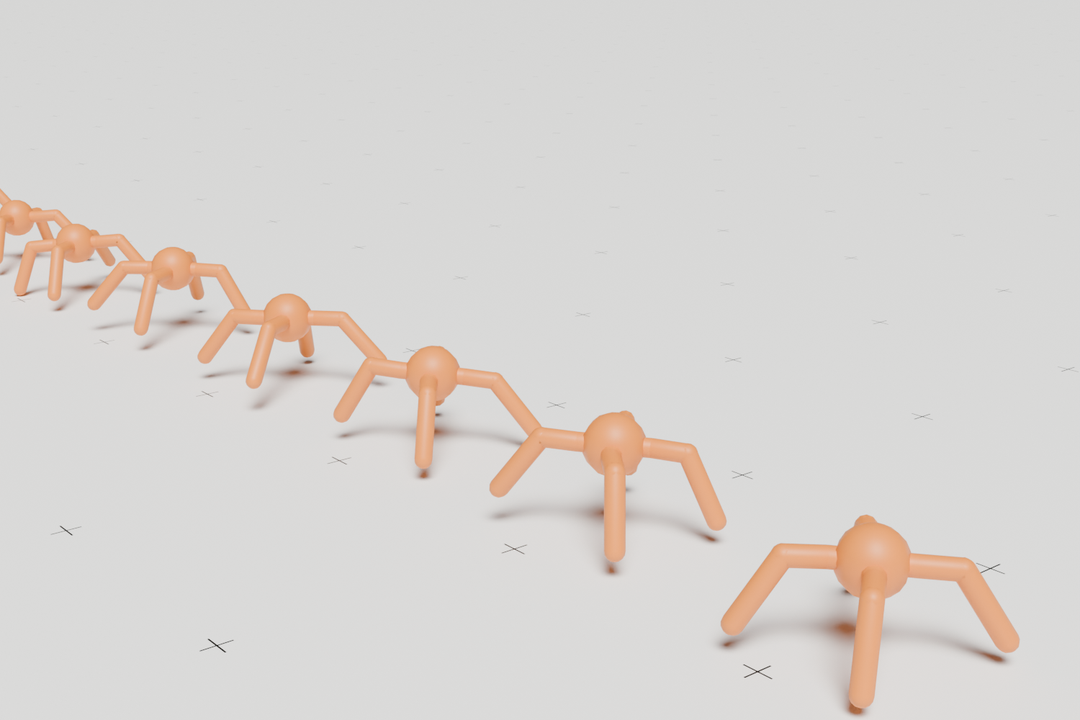}} &
    \subcaptionbox{\myfont\normalsize{AntRun}}{\includegraphics[width=0.29\columnwidth]{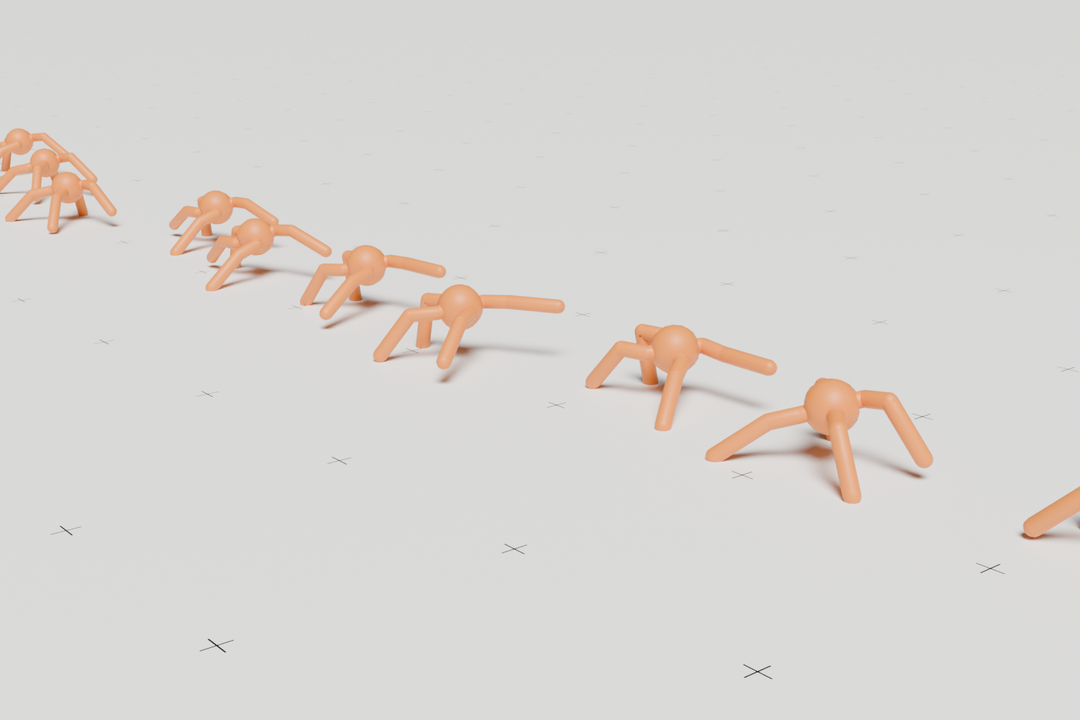}} &
    \subcaptionbox{\myfont\normalsize{}}{\includegraphics[width=0.29\columnwidth]{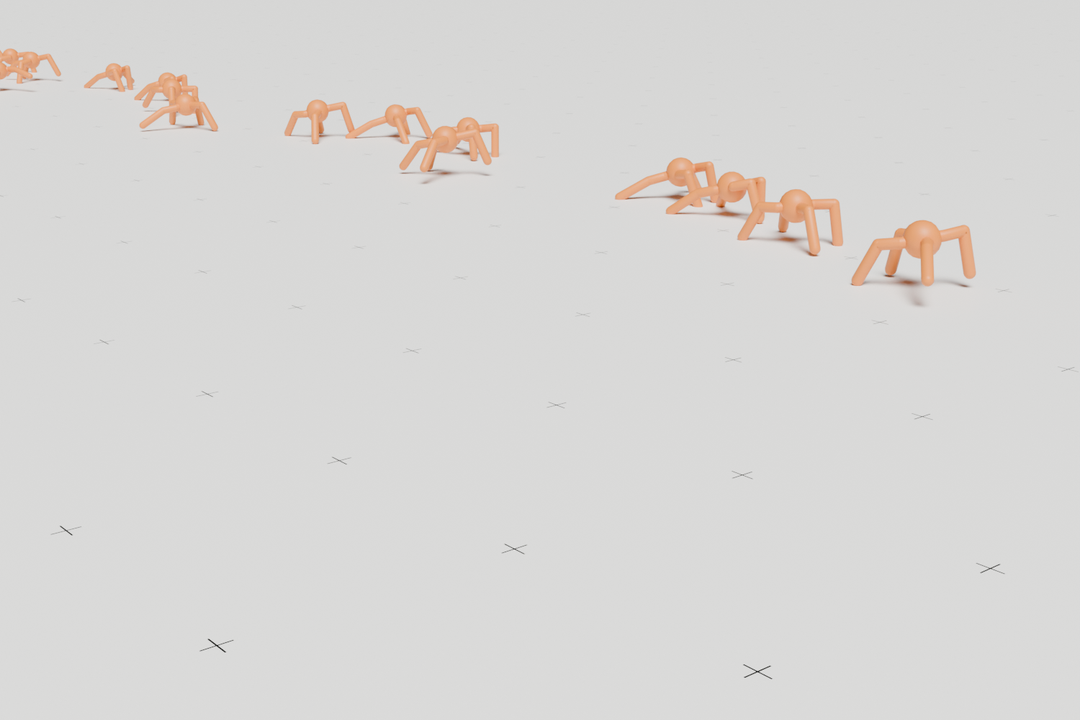}}
    \\
    \subcaptionbox{\myfont\normalsize{}}{\includegraphics[width=0.29\columnwidth]{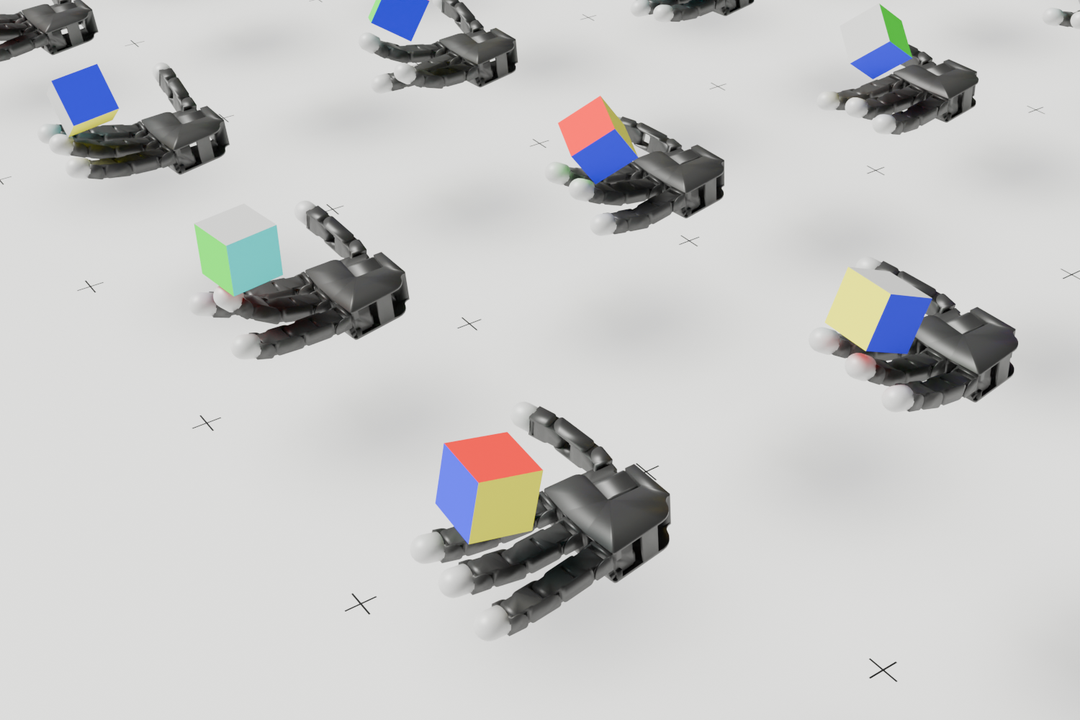}} &
    \subcaptionbox{\myfont\normalsize{HandReorient}}{\includegraphics[width=0.29\columnwidth]{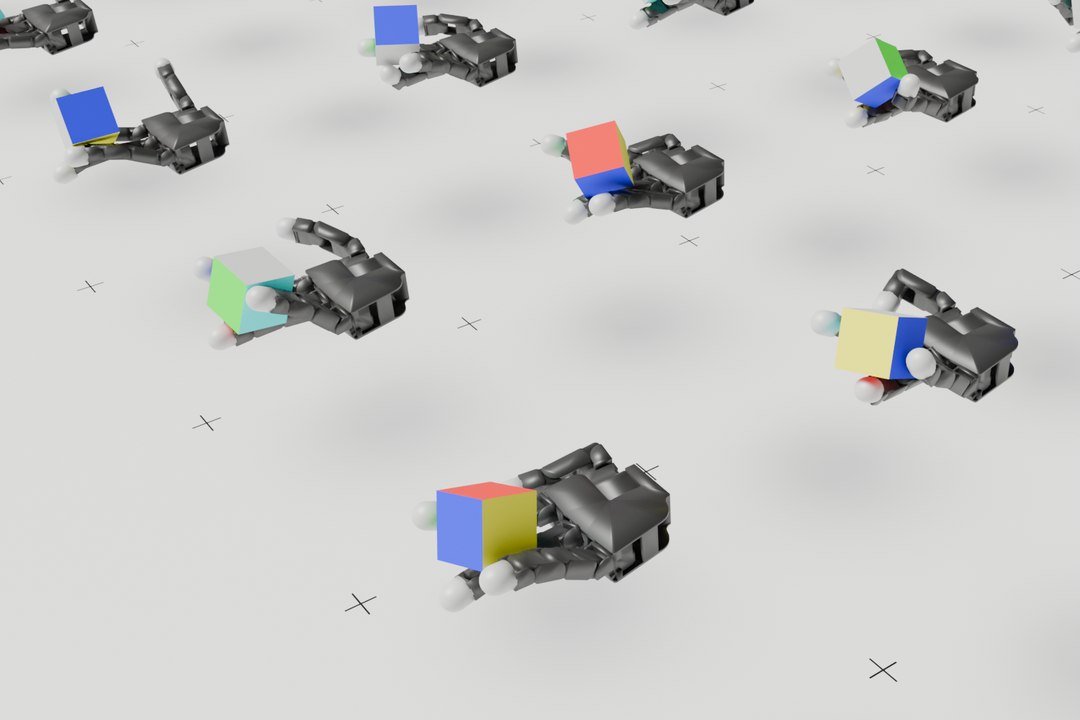}} &
    \subcaptionbox{\myfont\normalsize{}}{\includegraphics[width=0.29\columnwidth]{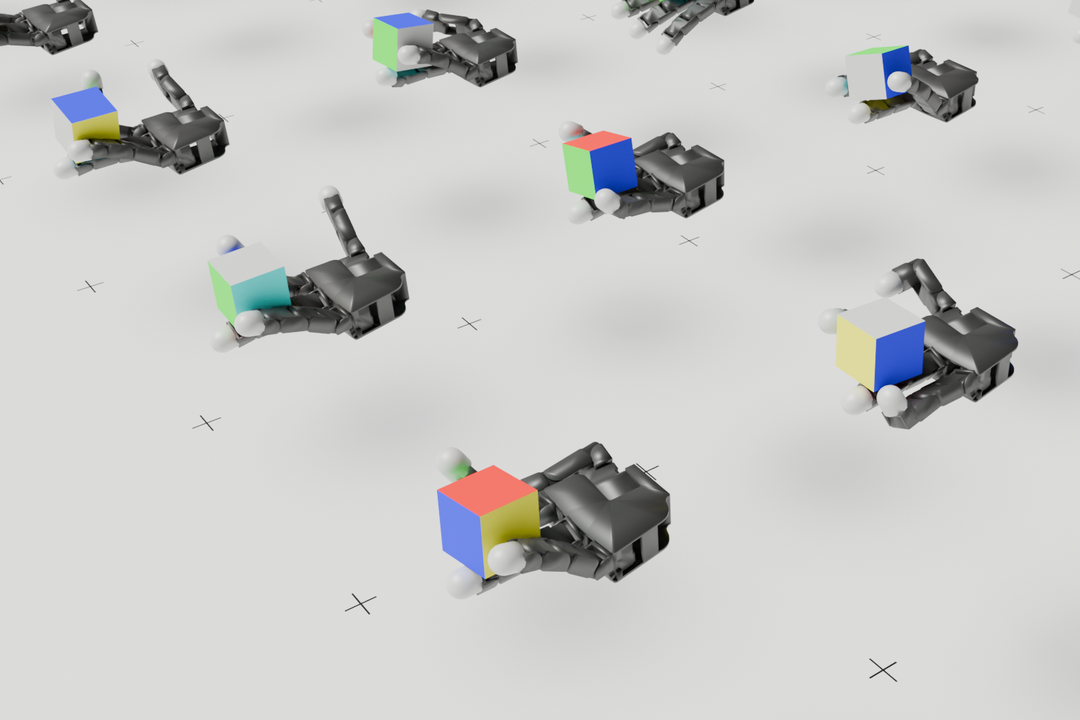}}
    \\
    \subcaptionbox{\myfont\normalsize{}}{\includegraphics[width=0.29\columnwidth]{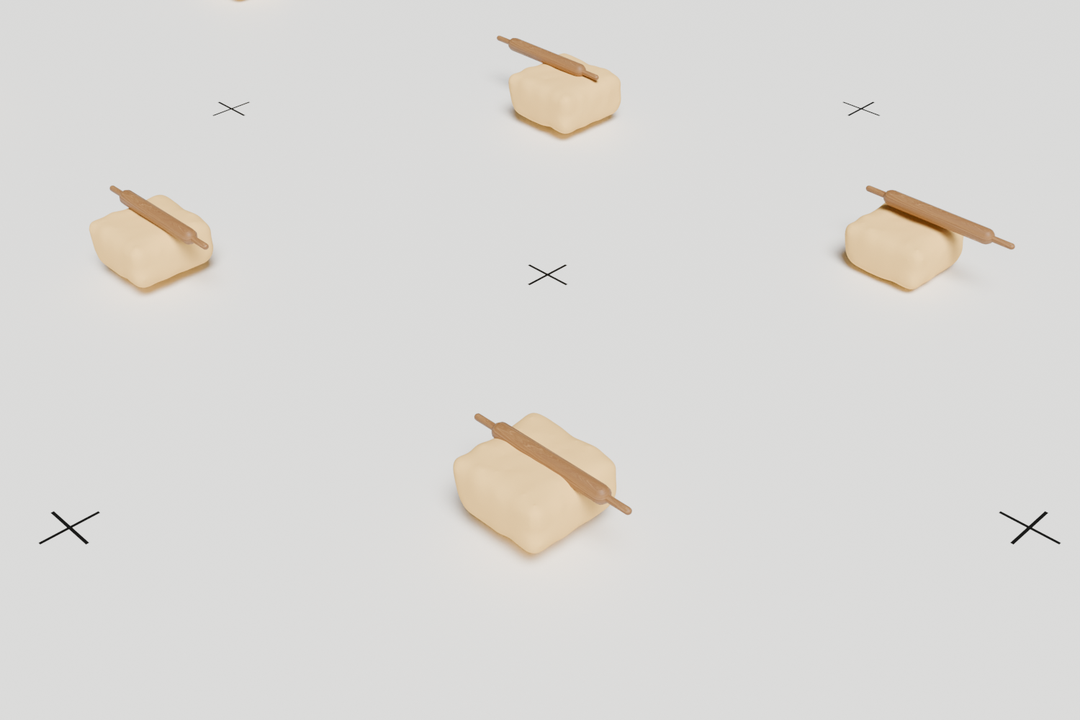}} &
    \subcaptionbox{\myfont\normalsize{RollingFlat}}{\includegraphics[width=0.29\columnwidth]{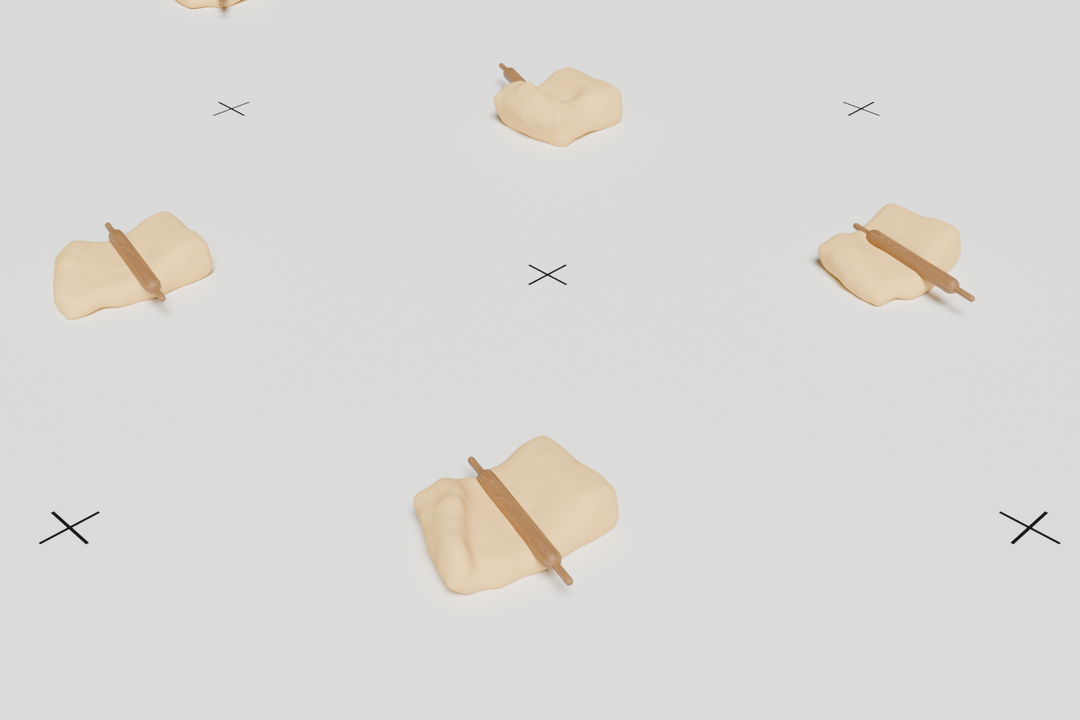}} &
    \subcaptionbox{\myfont\normalsize{}}{\includegraphics[width=0.29\columnwidth]{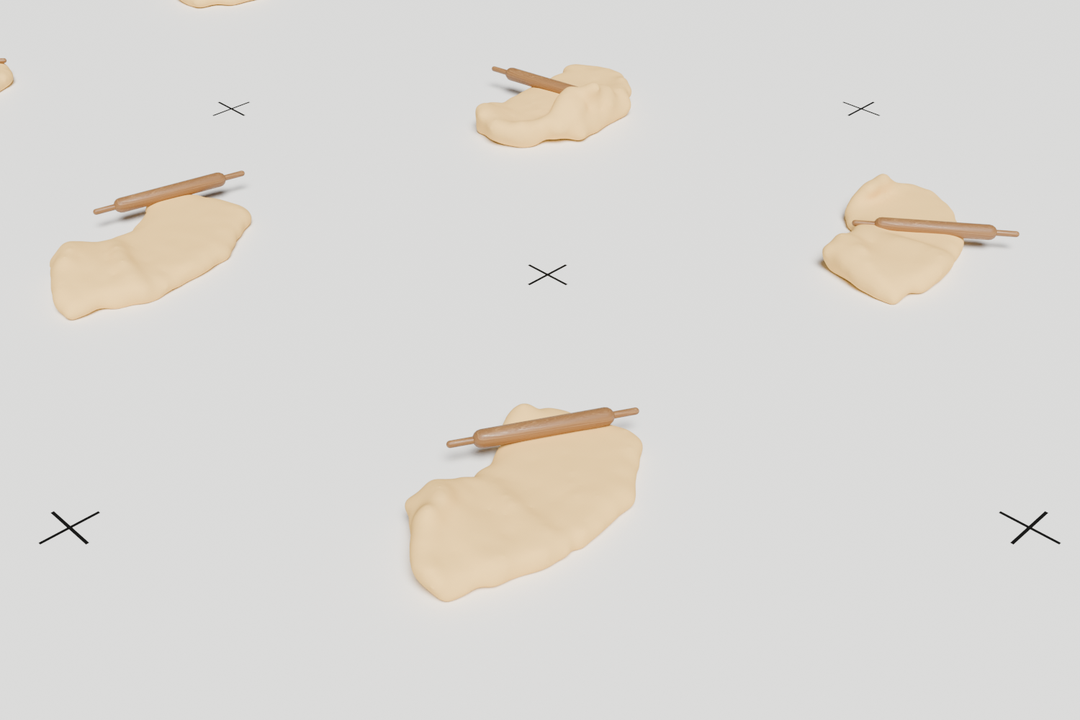}}
    \\
    \subcaptionbox{\myfont\normalsize{}}{\includegraphics[width=0.29\columnwidth]{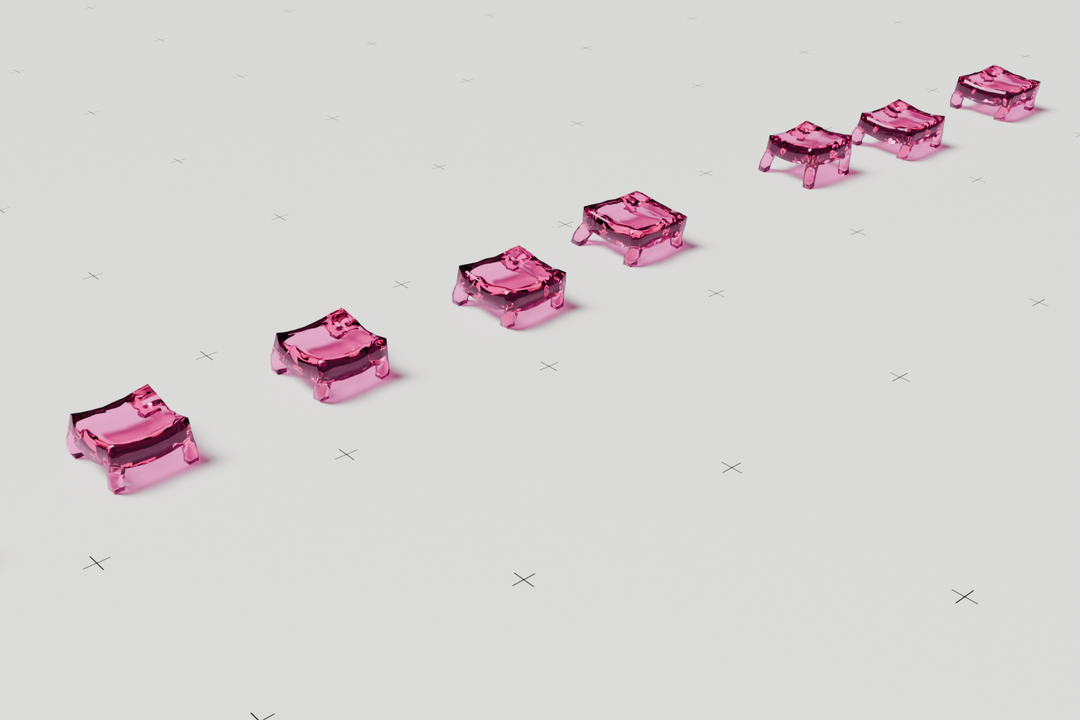}} &
    \subcaptionbox{\myfont\normalsize{SoftJumper}}{\includegraphics[width=0.29\columnwidth]{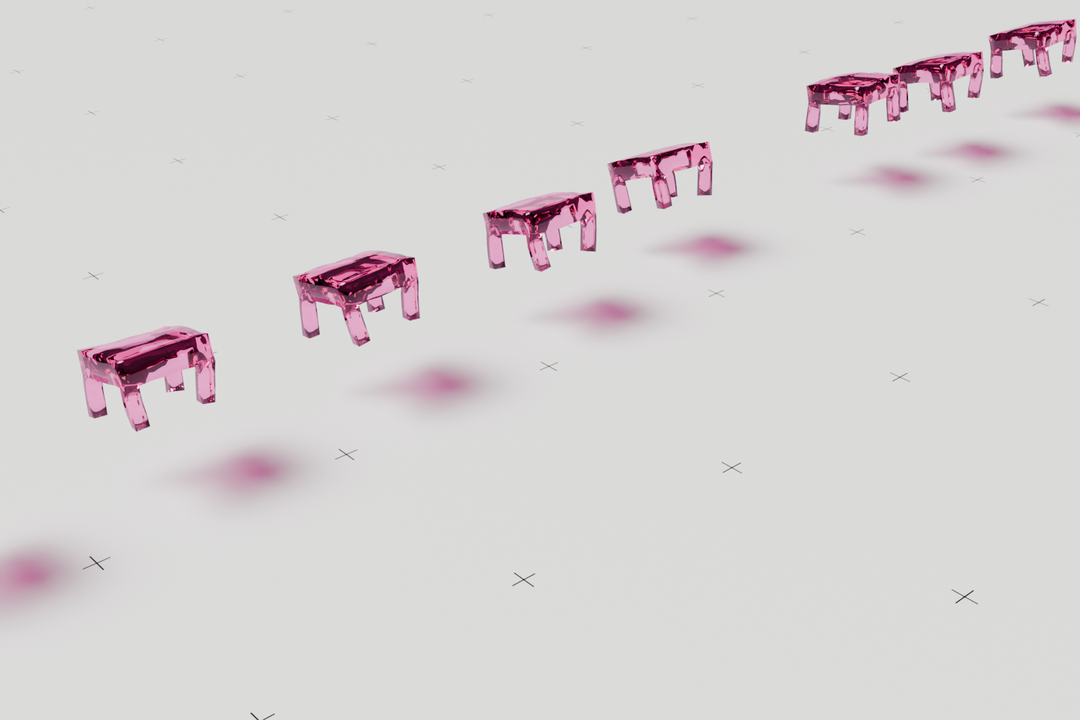}} &
    \subcaptionbox{\myfont\normalsize{}}{\includegraphics[width=0.29\columnwidth]{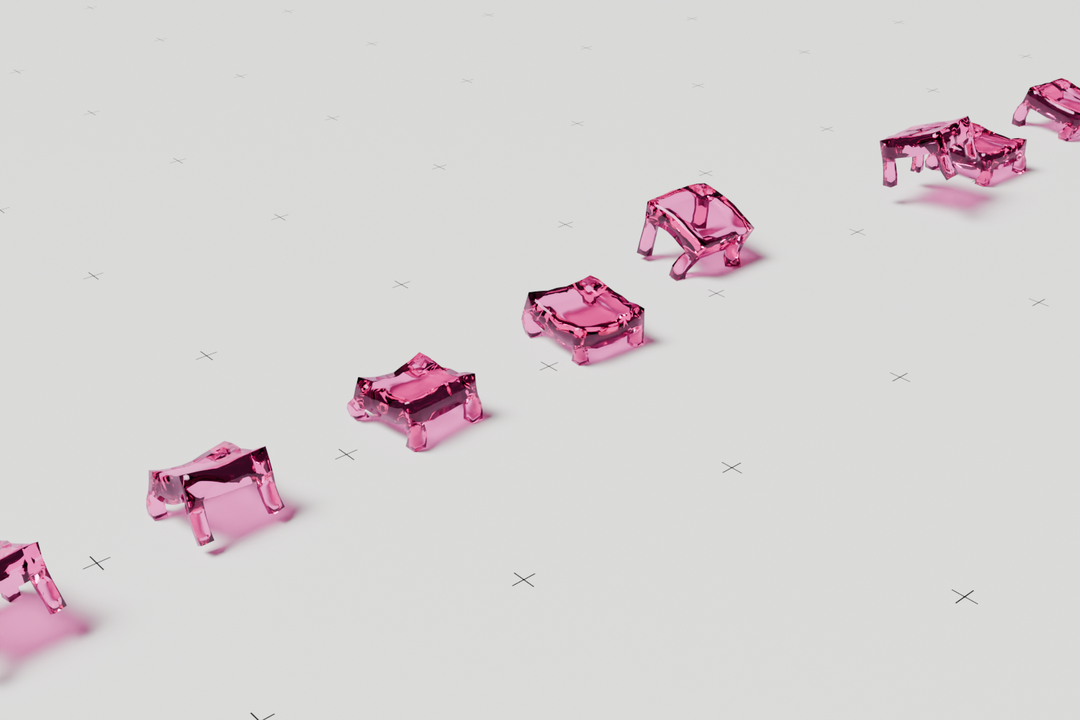}}
    \\
    \subcaptionbox{\myfont\normalsize{}}{\includegraphics[width=0.29\columnwidth]{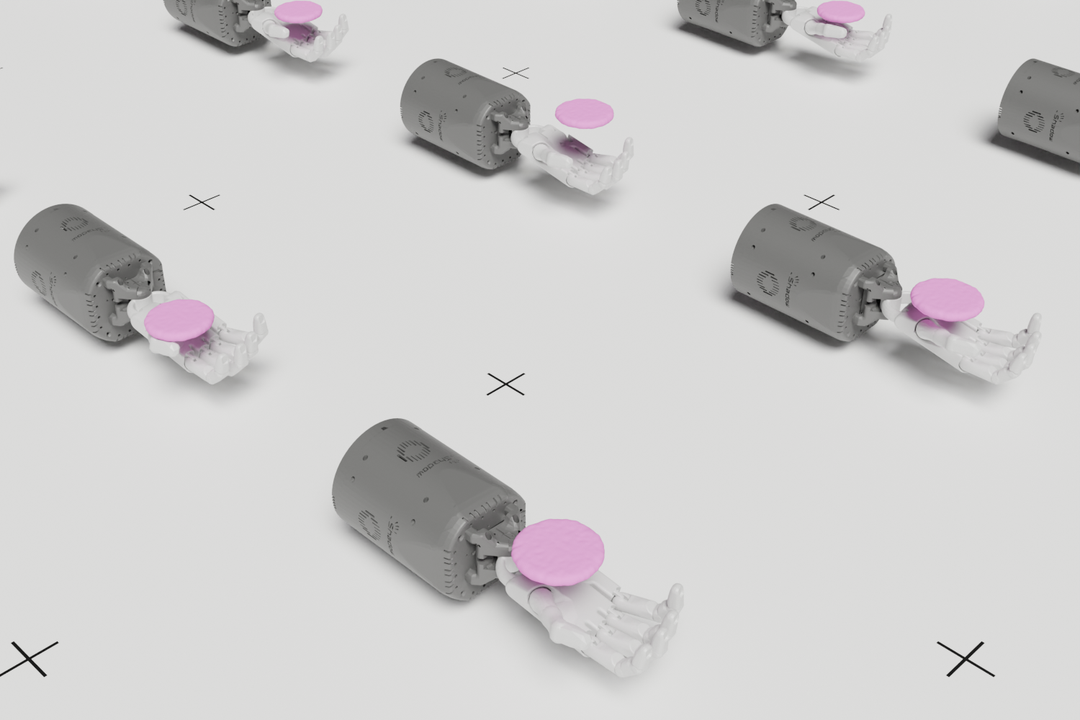}} &
    \subcaptionbox{\myfont\normalsize{HandFlip}}{\includegraphics[width=0.29\columnwidth]{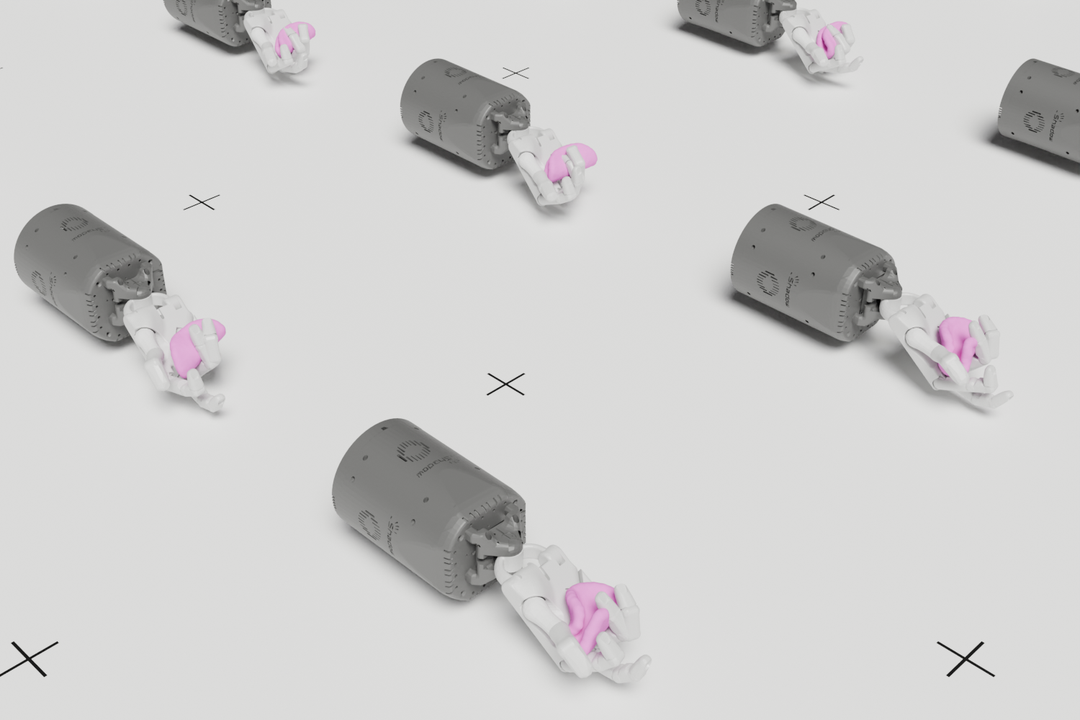}} &
    \subcaptionbox{\myfont\normalsize{}}{\includegraphics[width=0.29\columnwidth]{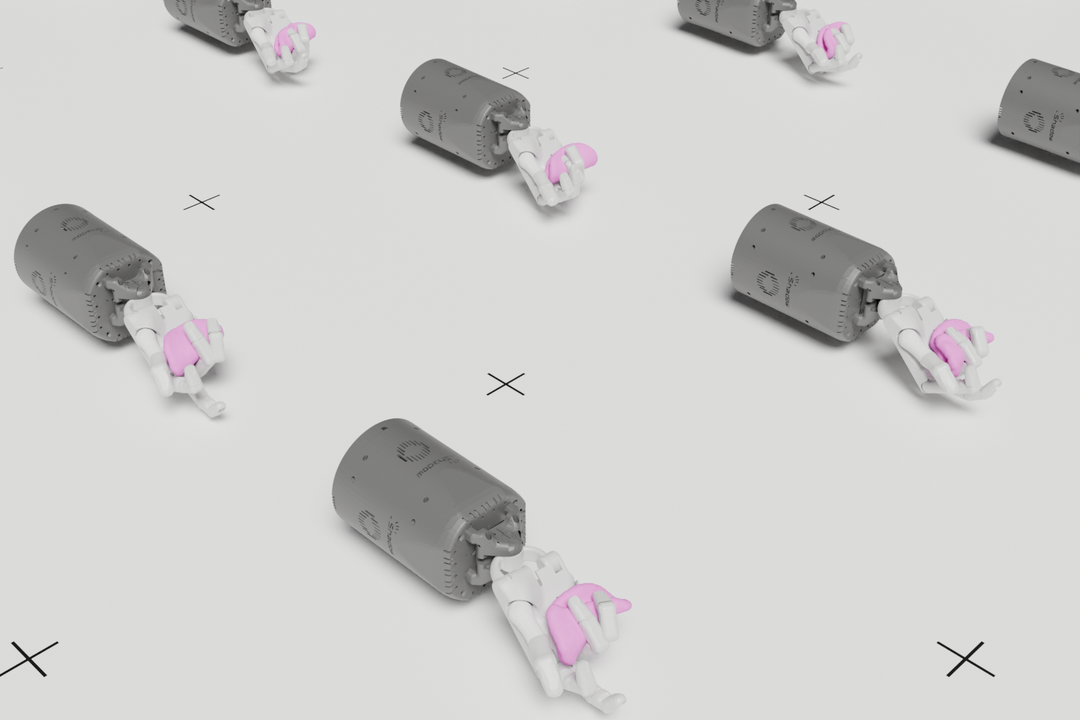}}
    \\
    \subcaptionbox{\myfont\normalsize{}}{\includegraphics[width=0.29\columnwidth]{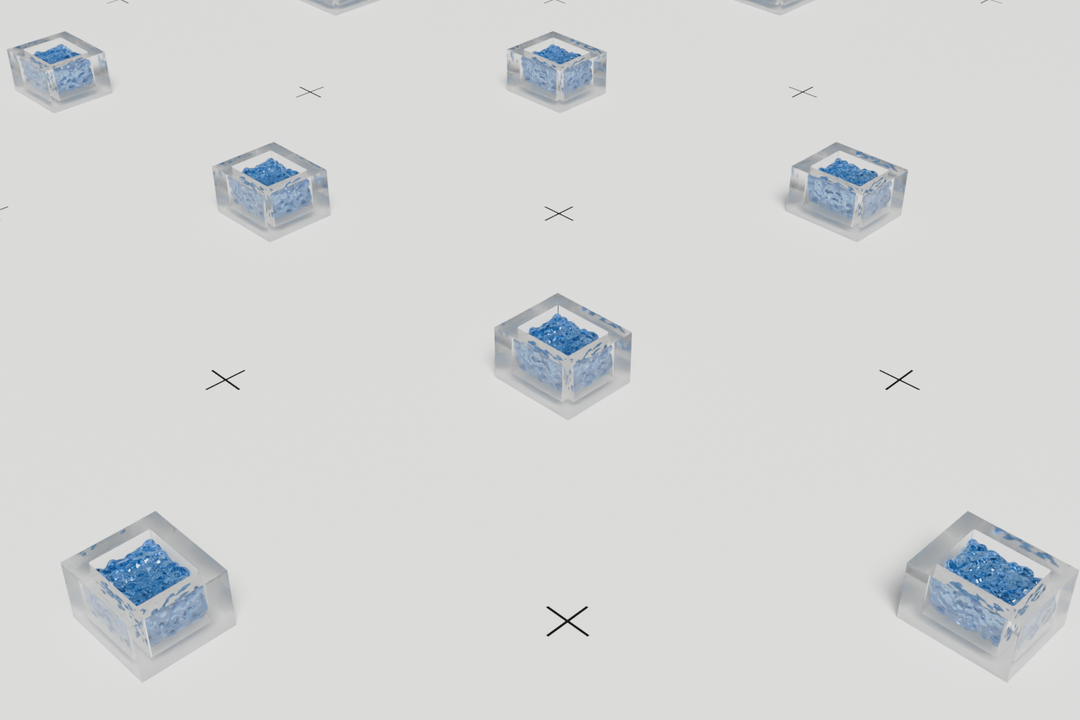}} &
    \subcaptionbox{\myfont\normalsize{FluidMove}}{\includegraphics[width=0.29\columnwidth]{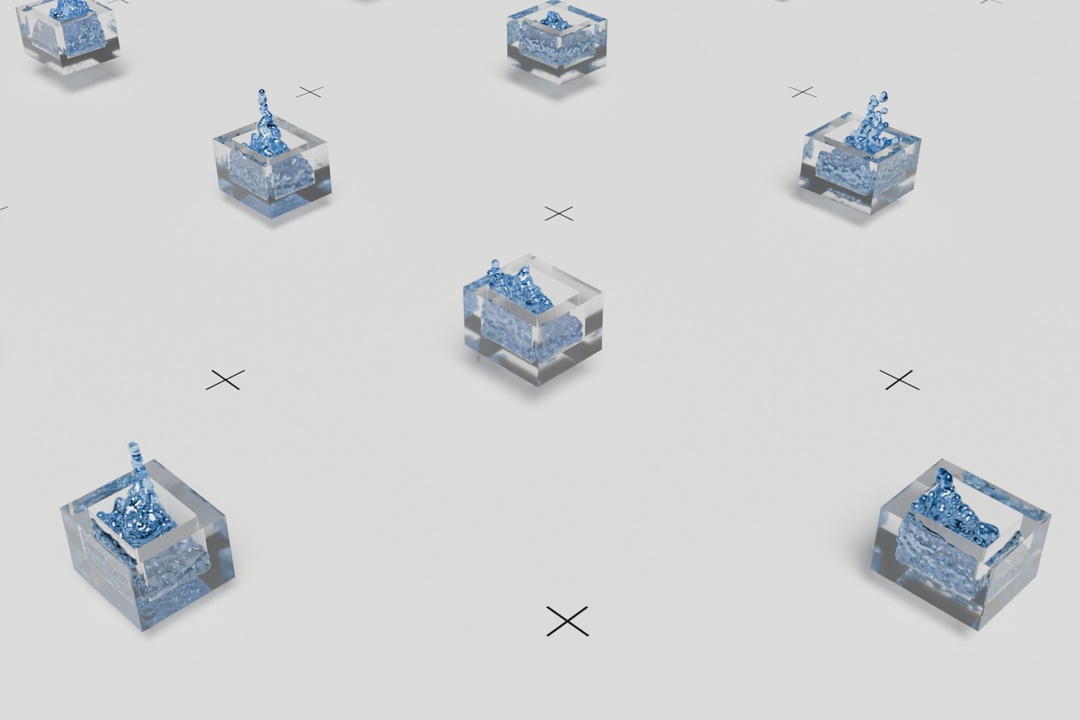}} &
    \subcaptionbox{\myfont\normalsize{}}{\includegraphics[width=0.29\columnwidth]{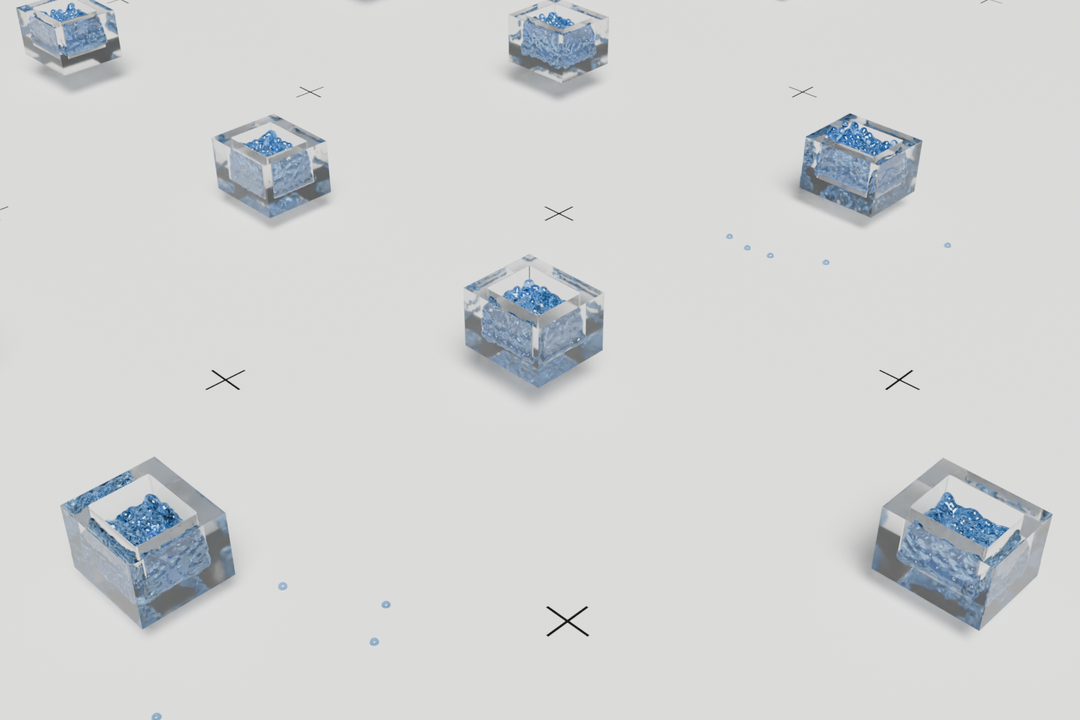}}
    \\[0.1em]
    & $\text{\myfont\normalsize{time}} \xrightarrow{\hspace*{3cm}}$ & \\
    \end{tabular}
    \caption{
        \textbf{Visualizations of trajectories from policies learned by \ouralgo in \oursim tasks.} The camera view is fixed between different time steps.
    }
    \label{fig:trajviz}
    \vspace{-3.0em}
\end{figure}

\clearpage
\subsection{Visualizations of optimization landscape}

\begin{figure}[H]
    \centering
    \captionsetup[subfigure]{labelformat=empty,skip=1em}
    \renewcommand\thesubfigure{}
    \captionsetup[subfigure]{position=below}
    \setlength{\tabcolsep}{0.1em}
    \def\arraystretch{3.5}
    \begin{tabular}{ccc}
    \subcaptionbox{\myfont\normalsize{APG}}{\includegraphics[width=0.32\columnwidth]{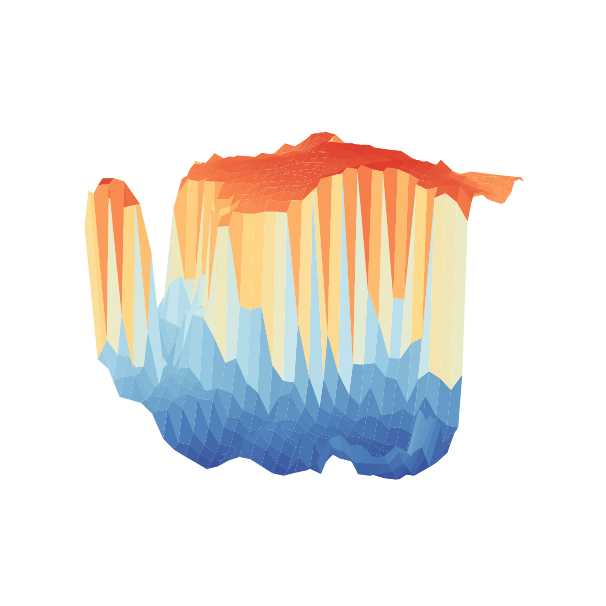}} &
    \subcaptionbox{\myfont\normalsize{SHAC}}{\includegraphics[width=0.32\columnwidth]{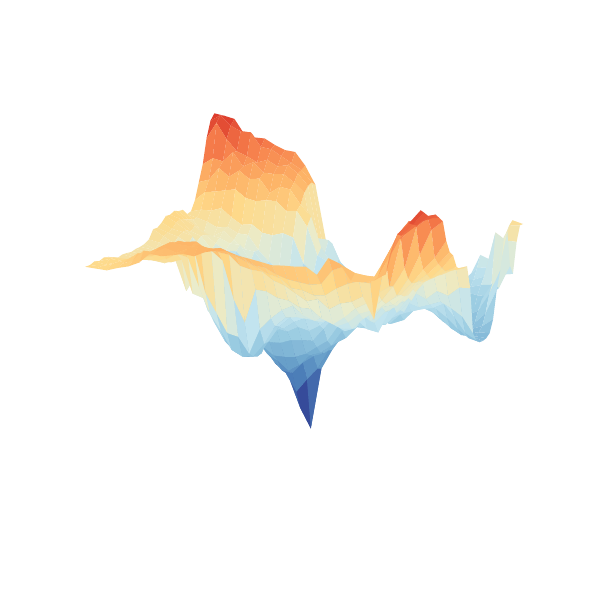}} &
    \subcaptionbox{\myfont\normalsize{\ouralgo}}{\includegraphics[width=0.32\columnwidth]{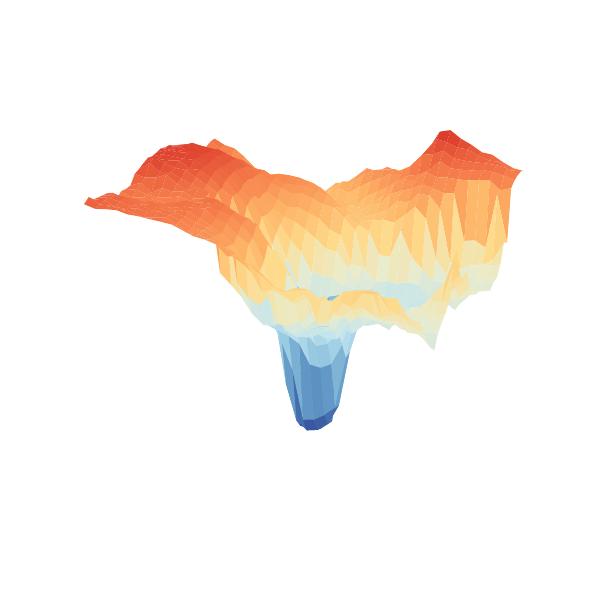}}
    \end{tabular}
    \caption{
        \textbf{Loss surface comparison between algorithms -- DFlex Ant.}
    }
    \label{fig:ll_ant}
\end{figure}

\begin{figure}[H]
    \centering
    \captionsetup[subfigure]{labelformat=empty,skip=2pt}
    \renewcommand\thesubfigure{}
    \captionsetup[subfigure]{position=below}
    \setlength{\tabcolsep}{0.1em}
    \def\arraystretch{3.5}
    \begin{tabular}{ccc}
    \subcaptionbox{\myfont\normalsize{APG}}{\includegraphics[width=0.32\columnwidth]{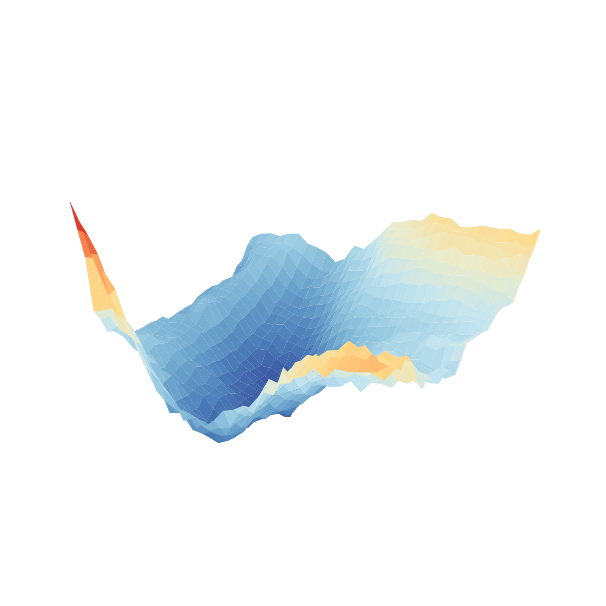}} &
    \subcaptionbox{\myfont\normalsize{SHAC}}{\includegraphics[width=0.32\columnwidth]{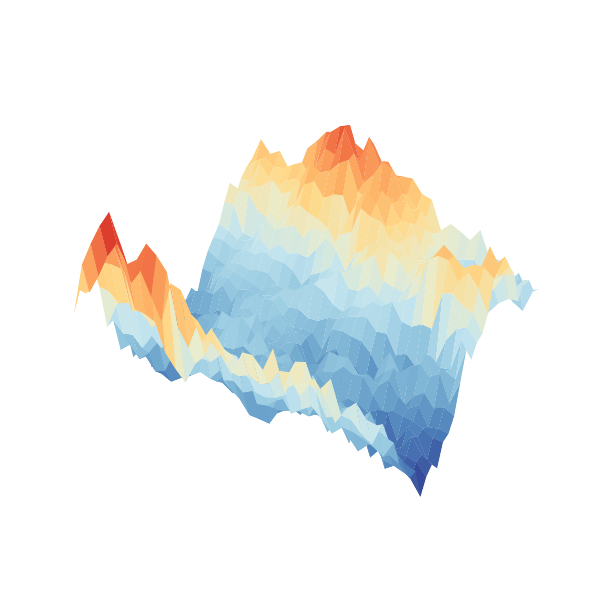}} &
    \subcaptionbox{\myfont\normalsize{\ouralgo}}{\includegraphics[width=0.32\columnwidth]{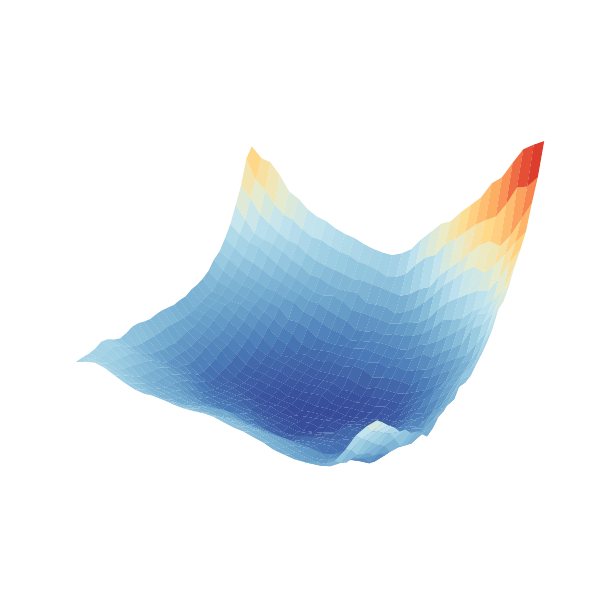}}
    \end{tabular}
    \caption{
        \textbf{Loss surface comparison between algorithms -- \oursim HandFlip.} 
    }
    \label{fig:ll_handflip}
\end{figure}

Following~\citep{xu2021accelerated}, we visualize loss surfaces to compare different algorithms and analyze the smoothness of the optimization landscape. We use undiscounted episodic returns for the loss values, and apply a symlog scale for visualization. We sample two filter-normalized random directions in policy parameter space and evaluate perturbations to final policies after training. We compare algorithms trained on DFlex Ant (Figure~\ref{fig:ll_handflip}) and \oursim HandFlip (Figure~\ref{fig:ll_ant}), observing that loss surfaces for \ouralgo appear to be flatter and smoother, compared to those of APG or SHAC.

\end{document}

%% file: math_commands.tex
\usepackage{dsfont}
\usepackage{amsmath,amsfonts,bm}

\def\eqref#1{equation~\ref{#1}}

\def\1{\bm{1}}

\DeclareMathAlphabet{\mathsfit}{\encodingdefault}{\sfdefault}{m}{sl}
\SetMathAlphabet{\mathsfit}{bold}{\encodingdefault}{\sfdefault}{bx}{n}

\newcommand{\Var}{\mathrm{Var}}

%% file: custom.tex
\usepackage[utf8]{inputenc} %
\usepackage[T1]{fontenc}    %
\usepackage{url}            %
\usepackage{booktabs}       %
\usepackage{amsfonts}       %
\usepackage{nicefrac}       %
\usepackage{microtype}      %
\usepackage{xcolor}         %
\usepackage{caption}        %

\usepackage{hyperref}
\hypersetup{
  colorlinks   = true, %
  urlcolor     = blue, %
  linkcolor    = blue, %
  citecolor    = blue %
}

\usepackage{multicol,multirow,array,enumerate,enumitem}
\usepackage{graphics,graphicx,tikz}
\usepackage{float,subcaption,wrapfig}
\usepackage{colortbl,tcolorbox,tabularx,makecell}
\usepackage{mathtools,amssymb,amsmath,amsfonts,bm,algorithmic}
\usepackage{xspace,ifthen,ragged2e,stackengine,changepage,cuted}
\usepackage{titlesec,footmisc}
\usepackage[ruled,vlined]{algorithm2e}
\usepackage{amsthm,thmtools,thm-restate}

\usepackage{pgfplots}
\pgfplotsset{compat=1.17}

\definecolor{rowblue}{HTML}{BD8DE1}

\usepackage{twemojis}
\newcommand{\Y}{\twemoji{check mark}}
\newcommand{\N}{\twemoji{cross mark}}  %

\newcommand{\norm}[1]{\left\lVert#1\right\rVert}

\newcommand{\abs}[1]{\lvert #1 \rvert}
\newcommand{\Exp}{\mathbb{E}}

\usepackage{siunitx}
\newcommand{\oursim}{Rewarped\xspace}
\newcommand{\ouralgo}{SAPO\xspace}  %
\newcommand{\OURALGO}{Soft Analytic Policy Optimization\xspace}

\newcommand{\pagelink}{rewarped.github.io}

\newcommand*{\myfont}{\fontfamily{cmbr}\fontseries{sb}\selectfont\color[HTML]{444444}}